\documentclass[fleqn,10pt]{wlscirep}
\usepackage[utf8]{inputenc}
\usepackage[T1]{fontenc}
\usepackage{lineno}
\usepackage{rotating}
\usepackage{multirow}
\usepackage{threeparttable}
\usepackage{hyperref}
\usepackage{appendix}
\title{ULTra-AV: A Unified Longitudinal Trajectory Dataset for Automated Vehicle}

\author[1,$\dag$]{Hang Zhou}
\author[1,$\dag$,*]{Ke Ma}
\author[1]{Shixiao Liang}
\author[1,*]{Xiaopeng Li}
\author[2,*]{Xiaobo Qu}
\affil[1]{University of Wisconsin-Madison, Department of Civil and Environmental Engineering, Madison, WI, 53706, USA}
\affil[2]{School of Vehicle and Mobility, Tsinghua University, Beijing, 100084, China}

\affil[*]{corresponding author(s): 
Xiaopeng Li (xli2485@wisc.edu),
Xiaobo Qu (drxiaoboqu@gmail.com),
Ke Ma (kma62@wisc.edu)
}
\affil[$\dag$]{these authors contributed equally to this work}

\begin{abstract}

Automated Vehicles (AVs) promise significant advances in transportation. Critical to these improvements is understanding AVs' longitudinal behavior, relying heavily on real-world trajectory data. Existing open-source trajectory datasets of AV, however, often fall short in refinement, reliability, and completeness, hindering effective performance metrics analysis and model development. This study addresses these challenges by creating a Unified Longitudinal TRAjectory dataset for AVs (Ultra-AV) to analyze their microscopic longitudinal driving behaviors. This dataset compiles data from 13 distinct sources, encompassing various AV types, test sites, and experiment scenarios. We established a three-step data processing:  1. extraction of longitudinal trajectory data, 2. general data cleaning, and 3. data-specific cleaning to obtain the longitudinal trajectory data and car-following trajectory data. 
The validity of the processed data is affirmed through performance evaluations across safety, mobility, stability, and sustainability, along with an analysis of the relationships between variables in car-following models. Our work not only furnishes researchers with standardized data and metrics for longitudinal AV behavior studies but also sets guidelines for data collection and model development.

\end{abstract}
\begin{document}

\flushbottom
\maketitle
  %  Click the title above to edit the author information and abstract}

\thispagestyle{empty}
\section*{Background \& Summary}

The advent of Automated Vehicles (AVs) marks a revolutionary change in the realm of transportation \cite{Calvert2018}. Various stakeholders, including transportation agencies, policymakers, urban planners, the automotive industry, and customers are paying attention to the potential impact of AV on traffic flow. Numerous studies have established a definitive and quantifiable link between macro-level traffic flow and micro-level longitudinal driving behavior\cite{Jin2016, Kerner2018}. This connection underscores the importance of understanding microscopic longitudinal AV behaviors to fully grasp their broader impacts on traffic. There is a particular emphasis on car-following behavior, the critical component in longitudinal driving behavior and arguably the most fundamental element in traffic flow \cite{Jiang2015, Qu2017}. The key to studying and comprehending the car-following driving behavior of AVs lies in the availability of real-world trajectory data, which contain a sequence of spatial positions, velocities, and ground truth accelerations over time and thus provide invaluable insights into AV behavior \cite{chen2023follownet}. Access to such AV trajectory data is imperative for stakeholders to generate reliable insights informing policy, infrastructure development, management strategies, traffic solutions, and AV design.

Numerous studies indicated that car-following behavior essentially impacts road traffic performance including safety, mobility, stability, and sustainability \cite{Axelsson2017, Wang2016, Ubiergo2016, Thiemann2008, feng2021intelligent}. AV trajectory data can offer direct insights into the impact on traffic in terms of these performance metrics. Safety is a priority for many stakeholders. In car-following behavior, the principal focus involves assessing the likelihood and timing of a rear-end collision, considering the vehicle's relative position and speed with respect to its preceding vehicle. Regarding mobility metrics, AVs have the potential to alter the driving strategy and following distance, thereby affecting throughput and traffic flow efficiency. While adopting an aggressive driving strategy or maintaining shorter following distances might enhance efficiency, such approaches would compromise stability, leading to diminished comfort and rising safety hazards. AVs are also expected to reduce overall fuel consumption of road traffic, thereby contributing to the achievement of environmental sustainability for future transportation. By directly analyzing AV trajectory data, stakeholders can assess these performance metrics, and develop strategies that enhance the positive impacts of AVs while mitigating potential negative impacts. 

Although AV trajectory data can provide intuitive insights into the performance metrics of AV, a significant limitation arises from the limited scenarios in which this data is collected. For example, trajectory data may be collected within a specific speed range or when the AV followed a vehicle adhering to a predetermined path \cite{Shi2021}. Thus, performance metrics derived from such constrained scenarios might present a biased view, failing to capture corner cases. This drawback underscores another crucial role of AV trajectory data: the accurate calibration of robust models that can run in the mirror of real-world conditions. These accurately calibrated models enable the exploration of broader impacts on traffic through simulation, including examining AV driving behavior in corner cases \cite{feng2023dense}. Furthermore, the models facilitate prediction interactions between AVs and human-driven vehicles. By accurate simulation results, the stakeholders can lay the groundwork for decision-making and strategic planning in anticipation of the forthcoming mixed traffic. 

% AV trajectory data can also be utilized to identify specific characteristics in AVs. For example, it facilitates the differentiation of AVs based on various criteria such as automation levels, powertrain types, and distinct brands and models. This detailed classification can enhance our understanding of how different configurations influence driving behavior and overall vehicle performance in real-world conditions.  

Recently, a surge in perception datasets of AVs—gathered through cameras and Light Detection and Ranging (LiDAR) in AV, such as BDD100K \cite{yu2020bdd100k}, Argoverse \cite{chang2019argoverse, wilson2023argoverse}, Waymo perception \cite{sun2020scalability}, KITTI \cite{geiger2013vision}, nuScenes \cite{caesar2020nuscenes}, ONCE \cite{mao2021one}, and ZOD \cite{alibeigi2023zenseact} datasets. These perception datasets are primarily used to predict the motion states of surrounding vehicles of AV and address basic safety conditions in AVs. However, they fall short of capturing the complex driving behaviors of AVs. In stark contrast, the collection of AV trajectory data, which depends on Global Positioning System (GPS) and Inertial Measurement Units (IMU), remains exceedingly scarce despite its critical importance as previously discussed. This scarcity is largely due to automakers' reluctance to voluntarily share their trajectory data with their automated driving technology. The high costs associated with renting test sites and vehicles with automated driving technology also pose barriers for researchers to collect this essential data. 

Despite these challenges, researchers worldwide have published several trajectory datasets under varying conditions and of differing sizes, as outlined in Table \ref{tab:summary}. However, these datasets often fall short in terms of refinement, reliability, and completeness, which limits their utility for comprehensive and precise studies of car-following behavior\cite{zhou2024review}. Firstly, not all datasets are specifically designed to collect trajectory data; they may inadvertently include car-following trajectory data alongside extraneous information, such as data on lateral vehicles that do not influence AV's car-following behavior \cite{sun2020scalability, kesting2007general}. This necessitates a selective refinement process to isolate the relevant trajectory data. Also, some datasets only contain raw data, which may include outliers or anomalies resulting from measurement errors, thus compromising data reliability. Thirdly, these datasets occasionally lack crucial details (e.g., vehicle length), requiring researchers to make educated guesses to fill these gaps. In light of these mentioned facts, currently, there is no unified and well-processed trajectory dataset available that encompasses multiple AVs across diverse experimental conditions and scenarios. This absence hinders the feasibility of conducting comprehensive studies on the impact of AVs on transportation. 

To address this gap, this study proposes a unified approach to processing trajectory datasets of AVs by enhancing their refinement, reliability, and completeness. While it would be impractical to conduct empirical research on all available vehicles with automated driving technology, developing a dataset in systematic and structured processes that compile the results of experimental campaigns conducted by various global research teams—including the author's group, Connected \& Autonomous Transportation Systems Laboratory (CATS Lab)—can provide substantial insights into the longitudinal behavior of AVs. Similar to how standard datasets such as ImageNet \cite{deng2009imagenet}, KITTI \cite{geiger2013vision}, and NGSIM \cite{punzo2011assessment} have revolutionized their respective fields, this work aims to establish an open-source Unified Longitudinal TRAjectory dataset for AVs (Ultra-AV) for future longitudinal behavior research of AV. The Ultra-AV dataset will facilitate the analysis of data, the development of models, and the identification of characteristics that influence AVs' impact on transportation. 

This study has the following contributions: 

\begin{itemize}
    \item This study systematically reviewed open-source AV trajectory datasets and detailed their collection scenarios and conditions. 
    \item This study developed a unified trajectory data format that includes essential elements for car-following behavior analysis of AVs, such as the position, speed, and acceleration of both the following AV (FAV) and the lead vehicle (LV). 
    \item This study introduced a standardized trajectory data processing methodology that involves multiple steps to enhance the refinement, reliability, and completeness of the data.
    \item This study validated the processed unified trajectory dataset through three key approaches: data collection methods, analysis of performance metrics, and development of AV models.
\end{itemize}

\begin{figure}[ht]
\centering
\includegraphics[width=\linewidth]{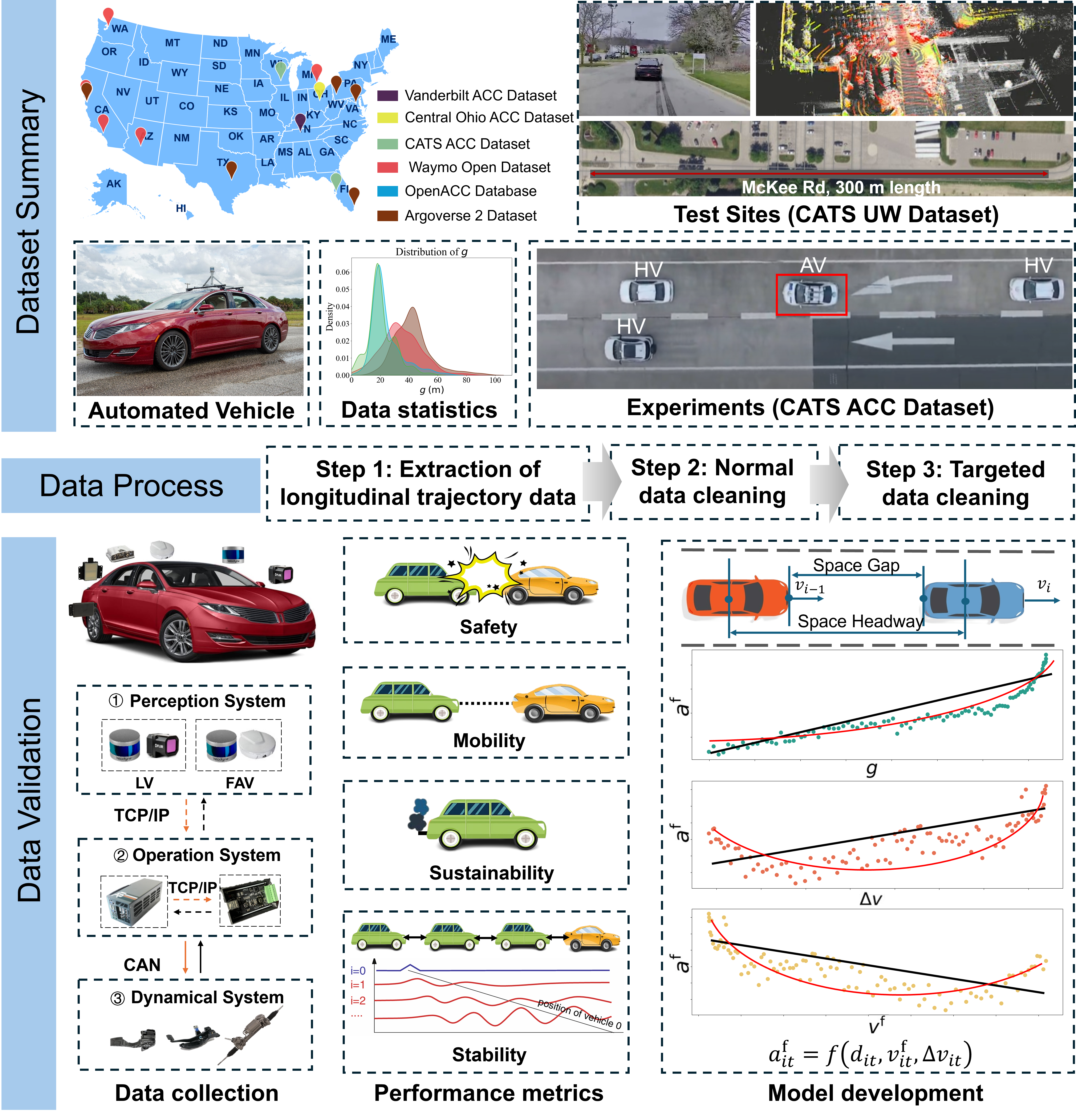}
\caption{A road map of this paper.}
\label{fig:workflow}
\end{figure}

To summarize, we leverage available open-source AV datasets to facilitate research. A comprehensive workflow for processing multiple open-source datasets to compile this dataset is illustrated in Figure \ref{fig:workflow}.   

\section*{Methods}

Efforts have been made globally to gather trajectory data related to AVs. We have examined 13 open-source datasets, each providing distinct insights into AV behavior across various driving conditions and scenarios. These open-source datasets are from six providers: %data provider
\begin{itemize}
    \item \textbf{Vanderbilt ACC Dataset}\cite{wang2019estimating}. Collected in Nashville, Tennessee by Vanderbilt University research group. Available at: \href{https://acc-dataset.github.io/datasets/}{https://acc-dataset.github.io/datasets/}.
    \item \textbf{CATS Open Datasets}\cite{shi2021empirical}. Three datasets were gathered in Tampa, Florida, and Madison, Wisconsin by the CATS Lab. Available at: \href{https://github.com/CATS-Lab}{https://github.com/CATS-Lab}.
    \item \textbf{OpenACC Database}\cite{makridis2021openacc}. Four datasets were collected across Italy, Sweden, and Hungary by the European Commission's Joint Research Centre. Available at: \href{https://data.europa.eu/data/datasets/9702c950-c80f-4d2f-982f-44d06ea0009f?locale=en}{https://data.europa.eu/data/datasets/9702c950-c80f-4d2f-982f-44d06ea0009f?locale=en}.
    \item \textbf{Central Ohio ACC Datasets}\cite{xia2023automated}. Two datasets were collated in Ohio by UCLA's Mobility Lab and Transportation Research Center. Available at: 
    
    \href{https://catalog.data.gov/dataset/advanced-driver-assistance-system-adas-equipped-single-vehicle-data-for-central-ohio}{https://catalog.data.gov/dataset/advanced-driver-assistance-system-adas-equipped-single-vehicle-data-for-central-ohio},
    
    \href{https://catalog.data.gov/dataset/advanced-driver-assistance-system-adas-equipped-two-vehicle-data-for-central-ohio}{https://catalog.data.gov/dataset/advanced-driver-assistance-system-adas-equipped-two-vehicle-data-for-central-ohio}.
    \item \textbf{Waymo Open Dataset}\cite{sun2020scalability, hu2022processing}. Two datasets were collected in six cities including San Francisco, Mountain View, and Los Angeles in California, Phoenix in Arizona, Detroit in Michigan, and Seattle in Washington by Waymo. Available at: \href{https://waymo.com/open/}{https://waymo.com/open/} and \href{https://data.mendeley.com/datasets/wfn2c3437n/2}{https://data.mendeley.com/datasets/wfn2c3437n/2}.
    \item \textbf{Argoverse 2 Motion Forecasting Dataset}\cite{wilson2023argoverse}. Collected from Austin in Texas, Detroit in Michigan, Miami in Florida, Pittsburgh in Pennsylvania, Palo Alto in California, and Washington, D.C. by Argo AI with researchers from Carnegie Mellon University and the Georgia Institute of Technology. Available at: \href{https://www.argoverse.org/av2.html}{https://www.argoverse.org/av2.html}.
\end{itemize}

The majority of the datasets reviewed involve AVs' long-time trajectories, which have been widely used in the analysis of AV behavior in the literature. However, the Waymo Open Dataset's Waymo Motion Dataset and the Argoverse 2 Motion Forecasting Dataset contain comparatively shorter trajectories, with durations of 9.1 seconds and 11 seconds at 10Hz, respectively. These datasets are primarily employed in research in motion forecasting. However, such datasets are typically collected in rural areas within complex traffic environments, which provide the opportunity to analyze AV behavior in challenging conditions. Consequently, this paper includes analyses of these two datasets. Other motion forecasting datasets always consist of even shorter trajectories, such as the Argoverse 1 Motion Forecasting Dataset\cite{chang2019argoverse} consisting of 5-second trajectories. Given that such a short duration may not adequately reflect AV behavioral patterns, our analysis does not consider these datasets.

\begin{figure}
    \centering
    \includegraphics[width=1\linewidth]{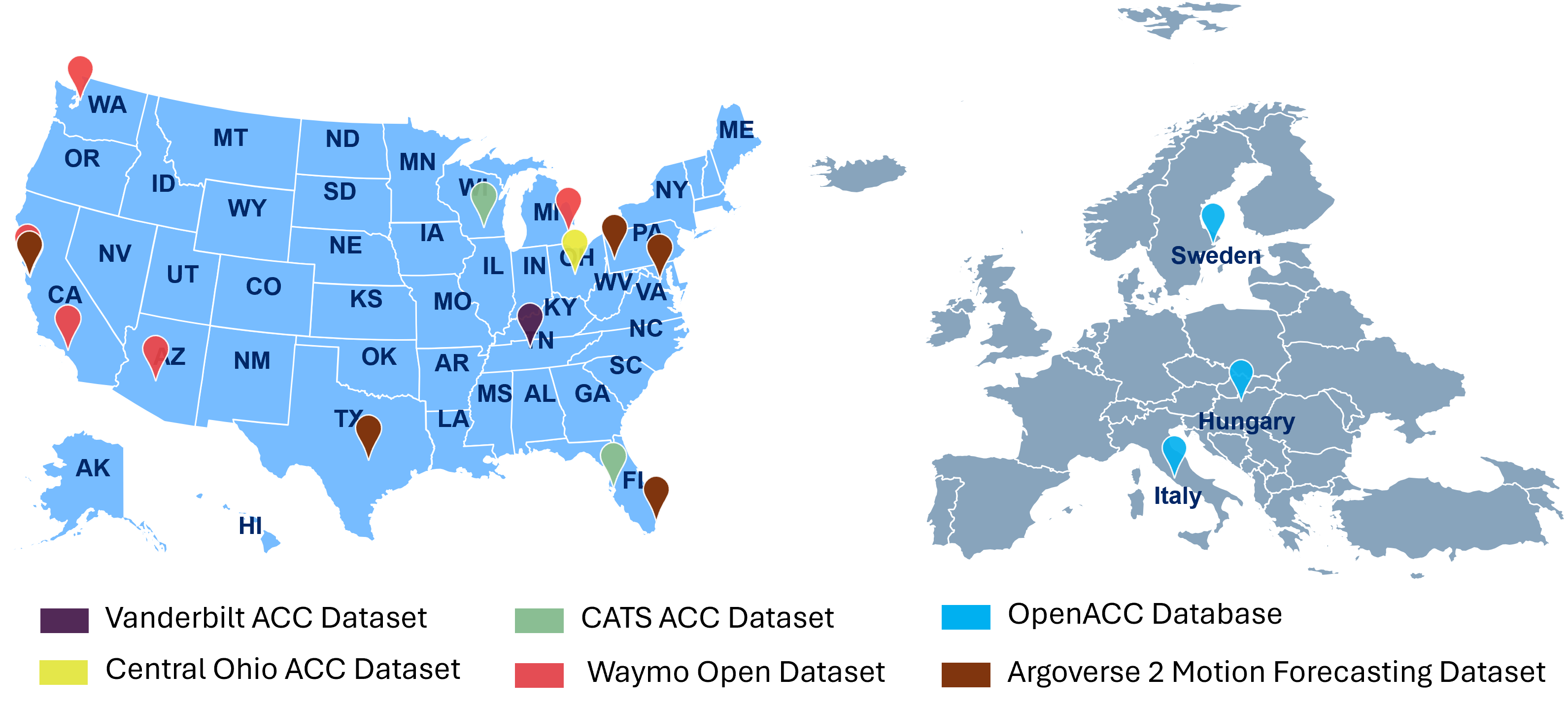}
    \caption{Test sites distribution of the trajectory datasets.}
    \label{fig:dataset}
\end{figure}

The locations of the test sites for the datasets are depicted in Figure \ref{fig:dataset}. The reviewed datasets are collected in several cities in the United States and Europe, which ensure the diversity and exemplarity among the selected cities \cite{xu2024unified}. Comprehensive details including data collection, AV information, test sites, and experiment settings are summarized in Table \ref{tab:summary}.

\begin{sidewaystable}[htbp]
  \footnotesize
  \centering
  \caption{Overview of the AV longitudinal trajectory open datasets.}
   \begin{threeparttable}
        \begin{tabular}{cp{5em}|p{4.3em}p{4.3em}p{4.3em}p{4.3em}p{4.3em}p{4.3em}p{4.3em}p{4.3em}p{4.3em}p{4.3em}p{4.3em}p{4.3em}p{4.3em}}
        \toprule
         \multicolumn{2}{c}{\textbf{Data Set\tnote{1}}} & \multicolumn{1}{c}{\textbf{1}} & \multicolumn{1}{c}{\textbf{2}} & \multicolumn{1}{c}{\textbf{3}} & \multicolumn{1}{c}{\textbf{4}} & \multicolumn{1}{c}{\textbf{5}} & \multicolumn{1}{c}{\textbf{6}} & \multicolumn{1}{c}{\textbf{7}} & \multicolumn{1}{c}{\textbf{8}} & \multicolumn{1}{c}{\textbf{9}} & \multicolumn{1}{c}{\textbf{10}} & \multicolumn{1}{c}{\textbf{11}} & \multicolumn{1}{c}{\textbf{12}} & \multicolumn{1}{c}{\textbf{13}} \\
        \midrule
          \multicolumn{1}{c}{\multirow{3}[4]{3.7em}{\textbf{Data Collection}}} & \textbf{Sensor} & Ublox\tnote{2} & Ublox C066-F9P & Ublox C066-F9P & LiDAR & Ublox 9 & Ublox 8 & RT-Range S\tnote{3} & \multirow{2}[1]{4.3em}{Diffusion of three types of sensors\tnote{4}} & \multirow{2}[1]{4.3em}{Diffusion of three types of sensors\tnote{5}} & \multirow{2}[1]{4.3em}{Diffusion of three types of sensors\tnote{5}}  & \multirow{2}[1]{4.3em}{Diffusion of two types of sensors\tnote{6}} & \multirow{2}[1]{4.3em}{Diffusion of two types of sensors\tnote{6}} & \multirow{2}[1]{4.3em}{Diffusion of three types of sensors\tnote{7}} \\
          & \textbf{Accurancy} & N/A   & 0.26m \& 0.089m/s & 0.26m \& 0.089m/s & N/A & 0.3 m \& 0.14 m/s  & N/A   & N/A &  &  &  &   &   & \\
          & \textbf{Frequency} & 10Hz  & 10Hz  & 1Hz   & 10Hz  & 10Hz  & 10Hz  & 10Hz  & 10Hz  & 10Hz  & 10Hz  & 10 Hz & 10 Hz & 10 Hz \\
          \cmidrule{1-2}    \multicolumn{1}{c}{\multirow{6}[2]{3.7em}{\textbf{AV Information}}} & \textbf{Level} & ADAS  & ADAS  & ADAS  & ADAS  & ADAS  & ADAS  & ADAS  & ADAS  & ADAS  & ADAS  & ADS   & ADS   & ADAS \\
          & \textbf{Brand} & N/A   & Lincoln & Lincoln & Lincoln & Hyundai & \multirow{5}[1]{4.3em}{7 vehicles, 6 brands, 7 models\tnote{8}} & \multirow{5}[1]{4.3em}{5 vehicles, 4 brands, 5 models\tnote{9}} & \multirow{5}[1]{4.3em}{12 vehicles, 9 brands, 12 models\tnote{10}} & Tesla   & \multirow{5}[1]{4.3em}{2 vehicles, 2 brands\tnote{11}}  & Waymo & Waymo & Ford \\
          & \textbf{Model} & N/A   & MKZ   & MKZ   & MKZ   & Ioniq hybrid &  &  & & N/A   &  & N/A   & N/A   & Fusion Hybrids\tnote{12} \\
          & \textbf{Year} & 2019 & 2016 \& 2017 & 2016 \& 2017 & 2016 & 2018 &  &  &  & N/A   &   & N/A   & N/A   & N/A \\
          & \textbf{Powertrain\tnote{13}} & N/A   & HV  & HV  & HV  & HV  &  &  & & N/A   &   & N/A   & N/A   & HV \\
          & \textbf{Deminsion} & SUV   & Sedan & Sedan & Sedan & Sedan &  &  & & Sedan   &   & SUV   & SUV   & Sedan \\
          \cmidrule{1-2}    \multicolumn{1}{c}{\multirow{3}[3]{3.7em}{\textbf{Test Sites}}} & \textbf{Accessibility} & public & public & public & public & public & public & closed & closed & public & public & public & public & public \\
          & \textbf{Road Type} & freeway & highway & highway & urban & highway & highway & rural & rural & freeway \& highway & freeway \& highway & urban & urban & urban \\
          & \textbf{Weather} & Ideal & Ideal & Ideal & Ideal & Ideal & Ideal & Ideal & Ideal & Ideal & Ideal & whole & whole & whole \\
          \cmidrule{1-2}    \multicolumn{1}{c}{\multirow{4}[20]{3.7em}{\textbf{Experiment Setting}}} & \textbf{Duration} & N/A   & 3 days & N/A   & 2024/03/01-2024/03/03 & 2020/10/27 & 2019/02/26-2019/02/28 & 2019/04/07, 2019/05/07 & 2019/10/06-2019/10/09 & 2021/09/1-2022/02/19 & 2022/03/15-2022/03/21 & N/A   & N/A   & N/A \\
          & \textbf{Speed Setting} & no    & [11.2,15.6], [20.1,24.6] & [22.4,24.6] & no    & no    & no    & straight: [25,27.8] \& curve: [13.9,16.7]  & [8.3,16.7] & no    & no    & no    & no    & no \\
          & \textbf{Headway Setting} & no    & 4 levels of headway settings & 4 levels of headway settings & no    & no    & shortest time headway setting & no    & 3 levels and mixed & no    & no    & no    & no    & no \\
          & \textbf{Drive type} & naturalistic drive & artificial drive & artificial drive & artificial drive & naturalistic drive & naturalistic drive & artificial drive & artificial drive & naturalistic drive & naturalistic drive & naturalistic drive & naturalistic drive & naturalistic drive \\
    \bottomrule
    \end{tabular}%
    \begin{tablenotes}
      \item[1] Datasets: 1 = Vanderbilt Two-vehicle ACC Dataset; 2 = CATS ACC Dataset; 3 = CATS Platoon Dataset; 4 = CATS UW Dataset; 5 = OpenACC Casale Dataset; 6 = OpenACC Vicolungo Dataset; 7 = OpenACC Asta Dataset; 8 = OpenACC ZalaZone Dataset; 9 = Ohio Single-vehicle Dataset; 10 = Ohio Two-vehicle Dataset; 11 = Waymo Perception Dataset; 12 = Waymo Motion Dataset; 13 = Argoverse 2 Motion Forecasting Dataset. Table \ref{tab:step1}-\ref{tab:correlation} also follows these IDs.
      \item[2] Ublox GNSS from Ublox company. (\href{https://www.u-blox.com/}{https://www.u-blox.com/})
      \item[3] The RT-Range S multiple target ADAS measurements solution by Oxford Technical Solutions Company. (\href{https://www.oxts.com/}{https://www.oxts.com/})
      \item[4] The three types of sensor are: Race Logic VBOX with 0.02 position accuracy and 0.03 m/s speed accuracy, Ublox 9 with 0.3 m position accuracy and 0.14 m/s speed accuracy, and a tracker App available from ZalaZone 10 m position accuracy and 0.28 m/s speed accuracy. 
      \item[5] For the Tesla vehicle, the three types of sensors are: a 32-line Velodyne LiDAR with 0.03m position accuracy, two pluggable USB monocameras, and an RT3000 with 0.01 m position accuracy from OXTS company. For the Ford vehicle, the RT3000 is replaced by the Novatel SPAN from Novatel Company (\href{https://novatel.com/}{https://novatel.com/}).
      \item[6] The two sensors are: LiDAR and the high-resolution pinhole camera.
      \item[7] The three types of sensors are: 32-line Velodyne LiDAR with 0.03 m position accuracy, high-resolution ring cameras, and front-view facing stereo cameras.
      \item[8] Ford S-Max 2018 ICEV SUV, KIA Niro 2019 HV SUV, Mini Cooper 2018 ICEV Hatchback, Mitsubishi Outlander PHEV 2018 HV SUV, Mitsubishi SpaceStar 2018 ICEV SUV, Peugeot 3008GTLine 2018 ICEV SUV, VW GolfE 2018 EV SUV.
      \item[9] Audi A6 2018 ICEV Sedan, Audi A8 2018 ICEV Sedan, BMW X5 2018 ICEV SUV, Mercedes AClass 2019 ICEV Sedan, Tesla Model3 2019 EV Sedan.
      \item[10] Audi A4 Avant 2019 HV SUV, Audi E-tron 2019 EV SUV, BMW I3S 2018 HV Hatchback, Jaguar I-Pace 2019 EV Hatchback, Mazda 3 2019 ICEV Sedan, Mercedes-Benz GLE 450 4Matic 2019 HV SUV, Smart BME Addv (developed by Budapest University of Technology and Economics), Skoda Octavia RS 2019 ICEV SUV, Tesla Model3 2019 EV Sedan, Tesla ModelS 2019 EV Sedan, Tesla ModelX 2016 EV Hatchback, Toyota RAV4 2019 HV SUV.
      \item[11] A retrofitted Tesla Sedan, and a retrofitted Ford Fusion Sedan from AutonomouStuff Company.
      \item[12] The Ford Fusion Hybrid is integrated with Argo AI self-driving technology.
      \item[13] ICEV/HV/EV: internal combustion engine vehicle/hybrid vehicle/electric vehicle.
    \end{tablenotes}
    \end{threeparttable}
  \label{tab:summary}%
\end{sidewaystable}

To enhance the refinement, reliability, and completeness of these datasets, this study proposes a three-step process to develop the Ultra-AV dataset by three steps: (1) extraction of longitudinal trajectory data; (2) general data cleaning to remove anomalies and errors; and (3) data-specific cleaning tailored for car-following behavior.  

\subsection*{Step 1: Extraction of longitudinal trajectory data}

The first step of the data process aims to obtain the unified longitudinal trajectory data. Thus, we identified and stored them with a unified data format. Before explaining the extraction process, we define the longitudinal trajectory used in this study. Define the index set $\mathcal{I}=\{1,...,I\}$ of longitudinal trajectories comprising a series of consecutive data points, where $I$ is the total number of trajectories. Each trajectory contains a series of consecutive time stamps $\mathcal{T}_i=\{t_{i0},t_{i1},...,t_{iT_i}\}$ with the same time gap $\Delta t$, where $T_i$ is the number of time stamps for trajectory $i$. Although the datasets we reviewed organize data in a similar "trajectory" format, they may contain different FAVs or LVs within different lanes in the same trajectory. To keep consistency, we define a longitudinal trajectory consisting of one FAV $c_i^{\mathrm{f}}$ to track the same LV $c^{\mathrm{l}}_i$ consistently in the same lane, without changing lanes throughout all time stamps in a trajectory $\mathcal{T}_i$. In a longitudinal trajectory $i$, the data point at time stamp $t$ corresponds to a state vector $\mathbf{s}_{it} = [a^{\mathrm{f}}_{it}, d_{it}, v^{\mathrm{f}}_{it}, \Delta v_{it}]$, where $a^{\mathrm{f}}_{it}$ denotes the longitudinal acceleration of $c^{\mathrm{f}}_i$, $d_{it}$ is the spatial gap (i.e., bumper-to-bumper distance) between $c^{\mathrm{l}}_i$ and $c^{\mathrm{f}}_i$, $v^{\mathrm{f}}_{it}$ is the velocity of $c^{\mathrm{f}}_i$, and $\Delta v_{it}$ is the velocity difference between $c^{\mathrm{l}}_i$ and $c^{\mathrm{f}}_i$.

Extracting longitudinal trajectory set $\mathcal{I}$ necessitates identifying the $c^{\mathrm{l}}_i$ and $c^{\mathrm{f}}_i$. The dataset can be categorized into two types by the identification procedures. In the first category, exemplified by the Vanderbilt ACC Dataset and CATS Open Datasets, the relationship of $c^{\mathrm{l}}_i$ and $c^{\mathrm{f}}_i$ is labeled in the dataset. The second category, including the Central Ohio Datasets, Waymo Open Dataset, and Argoverse 2 Motion Forecasting Dataset, provides information on all surrounding vehicles but does not specifically label the relationship. Thus, this paper proposes a unified algorithm to identify LVs among mass trajectories. The identification algorithm is shown as follows:
\begin{enumerate}
    \item Segment trajectories to exhibit AV's lane-changing behaviors. For the Central Ohio Dataset, which includes the lane ID where the vehicles are located, processing is straightforward. We segment the trajectories into multiple consecutive trajectories. Thus, each segmented trajectory maintains a consistent lane ID throughout its duration. For datasets that only offer vehicles' positions without specific lane IDs, such as the Waymo Motion Dataset and the Argoverse 2 Motion Forecasting Dataset, identification of consistent lane trajectories poses a challenge. To address this, we employ linear regression to identify trajectories that exhibit straight-driving behaviors, indicative of consistent lane ID. Here, we denote the set of trajectories from the original dataset as $\mathcal{J}^0$ to differentiate the original trajectory set from the set of longitudinal trajectories $I$ obtained after the processing. Besides, the Euler coordinate position of the center of mass of vehicle $k\in \mathcal{K}_j$ in trajectory $j\in \mathcal{J}^1$ at timestamp $t$ as $p^{\mathrm{x}}_{jtk}$ and $p^{\mathrm{y}}_{jtk})$, respectively, where $k=0$ represents the FAV and $k\in \mathcal{K}_j/\{0\}$ represents the surrounding vehicle. For trajectory $j\in \mathcal{J}^0$, apply the least squares method to fit a linear model with $\{p^{\mathrm{x}}_{jt0}\}_{t\in \mathcal{T}_j}$ as inputs and $\{p^{\mathrm{y}}_{jt0}\}_{t\in \mathcal{T}_j}$ as outputs. We then compute the R-squared ($R^2$) of the linear model of the trajectory $j$. Trajectories whose $R^2$ is less than a threshold are considered not a straight line. The threshold is set as 0.9 as determined from our preliminary experimental results.   Finally, these trajectories are excluded from set $\mathcal{J}^0$. The new trajectory set is denoted as $\mathcal{J}^1$.
    \item Identify preceding vehicles of the FAV. For each trajectory $j \in \mathcal{J}^1$ and time stamp $t \in \mathcal{T}_j$, we define the set of vehicles $\bar{\mathcal{K}}_{jt}$ where each vehicle $k \in \bar{\mathcal{K}}_{jt}$ must meet the following criteria: 1. $k \in \mathcal{K}_j \setminus \{0\}$, 2. $k$ is located in the same lane with the FAV, and 3. $k$ is located in front of the FAV. For the Central Ohio Dataset, which provides both the lane ID and Frenet coordinates \cite{xia2023automated} of surrounding vehicles, the identification of $\bar{\mathcal{K}}_{jt}$ is straightforward. However, datasets such as the Waymo Motion Dataset and the Argoverse 2 Motion Forecasting Dataset primarily consist of trajectories represented in Euler coordinates. To process this data, we first excluded vehicles not moving in the same direction as the FAV by removing any vehicle $k$ where the dot product $\mathbf{\sigma}_0 \cdot \mathbf{\sigma}_k$ is negative, where $\mathbf{\sigma}_{0t} = [p^{\mathrm{x}}_{j(t-1)0} - p^{\mathrm{x}}_{jt0}, p^{\mathrm{y}}_{j(t-1)0} - p^{\mathrm{y}}_{jt0}]$ and $\mathbf{\sigma}_{kt} = [p^{\mathrm{x}}_{j(t-1)k} - p^{\mathrm{x}}_{jtk}, p^{\mathrm{y}}_{j(t-1)k} - p^{\mathrm{y}}_{jtk}]$ are the direction vectors of the FAV and vehicle $k$, respectively. Next, we define the direction vector from the FAV to vehicle $k$, $\mathbf{\sigma}_{0kt} = [p^{\mathrm{x}}_{jt0} - p^{\mathrm{x}}_{jtk}, p^{\mathrm{y}}_{jt0} - p^{\mathrm{y}}_{jtk}]$. We use the dot product of $\mathbf{\sigma}_{0t}$ and $\mathbf{\sigma}_{0kt}$ to verify alignment in direction, removing vehicles where this value is less than 0.984\footnote{Considering the average width of a mid-size car as 6 feet \cite{carSizes2023}, with lane widths at 10 feet \cite{nactoLaneWidth}, resulting in a maximum deviation of 4 feet. We assume the AV follows the 3-second rules \cite{drivesafe3second} and a minimum speed of 5 mph, which results in a minimum spatial gap of 22 feet. Thus the maximum angle $\theta = \arctan(0.182)$, $\cos(\theta) \approx 0.984$.}. This threshold ensures that only vehicles moving in a closely similar direction to the FAV are retained for further analysis. 
    \item Identify the LV. For each trajectory $j \in \mathcal{J}^1$ and time stamp $t \in \mathcal{T}_j$, calculate the spatial headway $h_{jtk}$ by Equation \eqref{equ:headway} for each vehicle $k \in \bar{\mathcal{K}}_{jt}$ relative to the FAV. $h_{jtk}$ is defined as the distance between the centers of FAV and the surrounding vehicle. 
    \begin{align}
      h_{jtk} = \sqrt{(p^{\mathrm{x}}_{jtk} - p^{\mathrm{x}}_{jt0})^2 + (p^{\mathrm{y}}_{jtk} - p^{\mathrm{y}}_{jt0})^2} \label{equ:headway}
    \end{align}
    The vehicle with the smallest $h_{jtk}$ is considered as the LV for that timestamp, $c^{\mathrm{l}}_{jt} = \arg \min_k h_{jtk}$. If a trajectory $j$ involves multiple LVs over time, divide it into several longitudinal trajectories, each consistent with a single LV. Collect these into set  $\mathcal{I}$.
    \item Enhance identification by the relationship between spatial headway and speed. Our preliminary experimental results identified several inaccuracies using previous methods. To enhance the algorithm, an additional step has been integrated into the trajectory processing workflow. For each trajectory $i \in \mathcal{I}$, we compare the change in spatial gap, $\Delta d_i = d_{iT_i} - d_{i0}$, to the change estimated from speed differences, $\Delta \hat{d}_i = \sum_{t \in \mathcal{T}_i} \Delta t \cdot \Delta v_{it}$. In a consistent car-following scenario, these two changes should align closely. Therefore, if the relative difference $\frac{|\Delta d_i - \Delta \hat{d}_i|}{\Delta d_i}$ exceeds a threshold of 0.2, which is based on the preliminary experimental results, trajectory $i$ is deemed inaccurate and removed from set $\mathcal{I}$. 
\end{enumerate}

Following the refinement of longitudinal trajectories, key labels relevant to analyzing FAV behaviors are extracted from the processed data and formatted consistently. The labels retained are listed in Table \ref{tab:labels}, where each label is described with its definition and calculation methods. Notably, a default value of 4.5 meters, corresponding to the average length of a mid-size vehicle  \cite{carSizes2023}, is assigned to vehicle length if it is not explicitly provided in the dataset. This standardization ensures uniformity in the data. 
% Most HVs will not appear in multiple trajectories, so we label all their IDs as -1

\begin{table}[htbp]
  \centering
  \caption{Labels for the uniformed data format.}
    \begin{tabular}{p{6em}p{19em}p{18em}p{2em}}
    \toprule
    Label & Description & Notations and formulation & Unit \\
    \midrule
    Trajectory\_ID & ID of the longitudinal trajectory. & $i\in \mathcal{I}$. & N/A \\
    Time\_Index & Common time stamp in one trajectory. & $t\in \mathcal{T}_i, i\in \mathcal{I}$. & $\mathrm{s}$ \\
    ID\_LV & LV ID. & $c^{\mathrm{l}}_i, i\in \mathcal{I}$. Label each FAV with a different ID and all HVs with -1. & N/A \\
    Pos\_LV & LV position in the Frenet coordinate. & $p^{\mathrm{l}}_{it}=p^{\mathrm{f}}_{it}+h_{it}, i\in \mathcal{I}, t\in \mathcal{T}_i$. & $\mathrm{m}$ \\
    Speed\_LV & LV speed. & $v^{\mathrm{l}}_{it}=\frac{p^{\mathrm{l}}_{i(t+1)}-p^{\mathrm{l}}_{it}}{\Delta t}, i\in \mathcal{I}, t\in \mathcal{T}_i$. & $\mathrm{m}/\mathrm{s}$ \\
    Acc\_LV & LV acceleration. & $a^{\mathrm{l}}_{it}=\frac{v^{\mathrm{l}}_{i(t+1)}-v^{\mathrm{l}}_{it}}{\Delta t}, i\in \mathcal{I}, t\in \mathcal{T}_i$. & $\mathrm{m}/\mathrm{s}^2$ \\
    ID\_FAV & FAV ID. & $c^{\mathrm{f}}_i, i\in \mathcal{I}$. Label each FAV with a different ID. & N/A \\
    Pos\_FAV & FAV position in the Frenet coordinate. & $p^{\mathrm{f}}_{it}=p^{\mathrm{f}}_{i(t-1)}+\Delta t \cdot v^{\mathrm{f}}_{it}, i\in \mathcal{I}, t\in \mathcal{T}_i$. & $\mathrm{m}$ \\
    Speed\_FAV & FAV speed. & $v^{\mathrm{f}}_{it}=\frac{p^{\mathrm{f}}_{i(t+1)}-p^{\mathrm{f}}_{it}}{\Delta t}, i\in \mathcal{I}, t\in \mathcal{T}_i$. & $\mathrm{m}/\mathrm{s}$ \\
    Acc\_FAV & FAV acceleration. & $a^{\mathrm{f}}_{it}=\frac{v^{\mathrm{f}}_{i(t+1)}-v^{\mathrm{f}}_{it}}{\Delta t}, i\in \mathcal{I}, t\in \mathcal{T}_i$. & $\mathrm{m}/\mathrm{s}^2$ \\
    Space\_Gap & Bump-to-bump distance between two vehicles. & $g_{it}=p^{\mathrm{l}}_{it}-p^{\mathrm{f}}_{it} - l^{\mathrm{f}}/2 -l^{\mathrm{l}}/2, i\in \mathcal{I}, t\in \mathcal{T}_i$, where $l^{\mathrm{f}}$ and $l^{\mathrm{f}}$ are the length of the LV and the FAV. & $\mathrm{m}$ \\
    Space\_Headway & Distance between the center of two vehicles. & $h_{it}=p^{\mathrm{l}}_{it}-p^{\mathrm{f}}_{it}, i\in \mathcal{I}, t\in \mathcal{T}_i$. & $\mathrm{m}$ \\
    % Time\_Headway & The time interval between two vehicles as they pass a specific position. & $\tau_{it}=\frac{h_{it}}{v^{\mathrm{f}}_{it}}, i\in \mathcal{I}, t\in \mathcal{T}_i$. & $\mathrm{s}$ \\
    Speed\_Diff & Speed difference of the two vehicles. & $\Delta v_{it}=v^{\mathrm{l}}_{it}-v^{\mathrm{f}}_{it}, i\in \mathcal{I}, t\in \mathcal{T}_i$. & $\mathrm{m}/\mathrm{s}$ \\
    \bottomrule
    \end{tabular}%
  \label{tab:labels}%
\end{table}%

\subsection*{Step 2: General data cleaning}

The second step focuses on enhancing the reliability of the trajectory dataset by cleaning it including removing outliers and inputting missing values. Raw datasets may include abnormal values due to sensor errors, such as accelerations exceeding 100 $\mathrm{m}/\mathrm{s}^2$. Such errors can become more pronounced when calculating additional metrics through differentiation. We define these abnormal values as outliers and remove them by excluding data outside $\eta$ standard deviations from the mean, denoted as $std$ and $mean$. The cleaning process includes the following procedures:
\begin{enumerate}
    \item Mark missing values and outliers. The mean and standard deviation for each label are calculated, excluding the data previously identified as missing. We then mark outliers that fall outside the range $[\text{mean} - \eta \cdot \text{std}, \text{mean} + \eta \cdot \text{std}]$. The calculation of the mean and standard deviation and the marking will be repeated iteratively without considering the marked outliers until all outliers are marked. The labels to be identified and their respective $\eta$ values are summarized in Table \ref{tab:clean}. We use a conservative $\eta$ to ensure this step removes only genuine outliers without affecting general data.
    \item Remove or input the marked data points. Based on experience, if a label includes ten consecutive marked data points, we remove all these points to maintain accuracy in trajectory analysis. However, if there are fewer than ten consecutive marked data points within a label, use linear interpolation to replace the marked data to minimize data loss. Note that this interpolation is only done within the same trajectory.
    \item Re-organize the trajectory ID. After removing some data points, certain trajectories may become discontinuous. To address this, we follow these steps to re-organize the "Trajectory\_ID" and "Time\_index" labels:
    1) Split any trajectories where "Time\_index" is discontinuous into multiple new trajectories, assigning each a new "Trajectory\_ID".
    2) Remove short trajectories that contain fewer than 70 data points. This threshold is based on preliminary experiments.
    3) Renumber the "Trajectory\_ID" and "Time\_index" columns to start from 0, ensuring a continuous sequence.
    4) Update the labels "Position\_LV" and "Position\_FAV" to reflect changes in trajectory segmentation. 
\end{enumerate}

After these cleaning procedures, we obtained the longitudinal trajectory dataset, which includes both free-flow and car-following trajectories. Given the critical importance of car-following behavior in the study of AV longitudinal behaviors, we specifically extracted the car-following trajectories in the next step for further analysis and validation.

\begin{table}[htbp]
  \centering
  \caption{Identified criteria for Step 2 and Step 3.}
    \begin{tabular}{llllll}
    \toprule
    Steps & $v^{\mathrm{f}}$ & $a^{\mathrm{f}}$ & $v^{\mathrm{l}}$ & $a^{\mathrm{l}}$ & $d$ \\
    \midrule
    Step 2 & N/A & $\eta=10$ & N/A & $\eta=10$ & $\eta=10$ \\
    Step 3 & $[0.1, +\infty)$ & $[-5, 5]$ & $[0.1, +\infty)$ & $[-5, 5]$ & $(0, 120]$ \\
    \bottomrule
    \end{tabular}%
  \label{tab:clean}%
\end{table}%

\subsection*{Step 3: Data-specific cleaning}

In this step, we define a hard margin for certain labels to identify car-following trajectories. Though the car-following concept is a broad consensus among researchers, there is no universally accepted definition \cite{liu2018relationship}. Thus, we proposed several thresholds derived from both a review of the relevant literature and empirical analysis of the data to identify car-following behavior, which are also summarized in Table \ref{tab:clean}:
\begin{itemize}
    \item A minimum speed threshold of 0.1 m/s, below which FAVs are considered to be stationary based on empirical observations.  
    \item A spatial distance threshold whereby an LV is situated within 120 meters on the same lane as the FAV, identifying it from free-flow traffic conditions \cite{mai2019advancement}.
    \item An acceleration range set of FAV is between -5 $\mathrm{m}/\mathrm{s}^2$ to 5 $\mathrm{m}/\mathrm{s}^2$ \cite{alotibi2020anomaly}.
\end{itemize}

The following process of this step is similar to Step 2. Data points that fall outside these established thresholds will be removed for exclusion or rectified through linear interpolation. After these three steps, we finally obtained the car-following trajectory dataset.

%The Methods should include detailed text describing any steps or procedures used in producing the data, including full descriptions of the experimental design, data acquisition assays, and any computational processing (e.g. normalization, image label extraction). See the detailed section in our submission guidelines for advice on writing a transparent and reproducible methods section. Related methods should be grouped under corresponding subheadings where possible, and methods should be described in enough detail to allow other researchers to interpret and repeat, if required, the full study. Specific data outputs should be explicitly referenced via data citation (see Data Records and Citing Data, below).

%Authors should cite previous descriptions of the methods under use, but ideally the method descriptions should be complete enough for others to understand and reproduce the methods and processing steps without referring to associated publications. There is no limit to the length of the Methods section. Subheadings should not be numbered.

\section*{Data Records}

The statistics for the dataset after each step of the processing mentioned in the last section are recorded in Table \ref{tab:step1}-\ref{tab:step3}. 

\begin{table}[htbp]
  \footnotesize
  \centering
  \caption{Statistical results of the data following Step 1 processing.}
        \begin{tabular}{llrrrrrrrrrrrrr}
    \toprule
    Label & Statistics & 1     & 2     & 3     & 4     & 5     & 6     & 7     & 8     & 9     & 10    & 11    & 12    & 13 \\
    \midrule
    \multirow{4}[2]{*}{$v^{\mathrm{l}}$} & mean  & 29.3  & 15.0  & 23.2  & 3.5   & 32.2  & 26.3  & 18.3  & 9.3   & 15.9  & 14.0  & 6.8   & 5.7   & 4.2 \\
          & std   & 1.5   & 8.9   & 0.7   & 0.5   & 6.6   & 9.5   & 4.7   & 4.0   & 11.0  & 9.9   & 7.3   & 6.2   & 4.7 \\
          & min   & 25.7  & 0.0   & 17.4  & 1.7   & 0.0   & 0.0   & 0.0   & -0.1  & 0.0   & 0.0   & -0.3  & 0.0   & 0.0 \\
          & max   & 34.8  & 26.7  & 24.7  & 4.0   & 40.1  & 40.8  & 33.8  & 27.5  & 42.6  & 39.8  & 28.5  & 32.9  & 22.9 \\
    \midrule
    \multirow{4}[2]{*}{$a^{\mathrm{l}}$} & mean  & 0.0   & 0.0   & 0.0   & 0.0   & 0.0   & 0.0   & 0.0   & 0.0   & 0.3   & 0.1   & 0.1   & 0.0   & 0.0 \\
          & std   & 0.3   & 3.0   & 0.2   & 0.4   & 0.6   & 1.9   & 0.8   & 0.7   & 3.5   & 2.7   & 0.7   & 9.3   & 7.4 \\
          & min   & -2.7  & -239.1 & -1.8  & 0.0   & -10.7 & -330.6 & -203.5 & -162.3 & -29.7 & -18.6 & -2.9  & -304.6 & -160.8 \\
          & max   & 3.8   & 112.9 & 5.8   & 10.0  & 13.2  & 167.3 & 165.6 & 160.6 & 92.9  & 90.5  & 3.5   & 298.6 & 161.4 \\
    \midrule
    \multirow{4}[2]{*}{$v^{\mathrm{f}}$} & mean  & 29.3  & 14.9  & 23.2  & 3.2   & 32.2  & 26.4  & 18.3  & 9.3   & 16.6  & 14.4  & 7.0   & 6.1   & 4.5 \\
          & std   & 1.7   & 9.0   & 0.9   & 0.7   & 6.8   & 9.6   & 4.9   & 4.1   & 10.6  & 9.3   & 7.1   & 6.3   & 4.9 \\
          & min   & 25.7  & 0.0   & 17.6  & 1.2   & 0.0   & 0.0   & 0.0   & 0.0   & 0.0   & 0.0   & -0.3  & -1.0  & 0.0 \\
          & max   & 34.9  & 28.6  & 25.4  & 4.2   & 41.2  & 42.8  & 34.5  & 27.5  & 35.6  & 31.9  & 29.6  & 58.2  & 24.0 \\
    \midrule
    \multirow{4}[2]{*}{$a^{\mathrm{f}}$} & mean  & 0.0   & 0.0   & 0.0   & 0.0   & 0.0   & 0.0   & 0.0   & 0.0   & 0.0   & 0.0   & 0.0   & 0.0   & 0.0 \\
          & std   & 0.2   & 3.2   & 0.3   & 0.1   & 0.6   & 2.0   & 0.8   & 0.7   & 0.6   & 0.6   & 0.7   & 10.1  & 6.8 \\
          & min   & -1.1  & -239.7 & -1.0  & -1.2  & -9.8  & -364.3 & -203.5 & -162.3 & -5.6  & -5.0  & -2.9  & -447.5 & -228.3 \\
          & max   & 0.6   & 135.8 & 6.2   & 1.0   & 10.1  & 167.3 & 157.0 & 160.6 & 6.1   & 3.3   & 2.9   & 423.3 & 219.2 \\
    \midrule
    \multirow{4}[2]{*}{$d$} & mean  & 36.2  & 29.3  & 37.3  & 13.8  & 42.6  & 37.1  & 25.3  & 20.5  & 25.5  & 22.7  & 20.5  & 13.3  & 14.0 \\
          & std   & 6.7   & 16.2  & 10.5  & 4.8   & 15.3  & 26.7  & 12.3  & 12.7  & 46.7  & 63.3  & 16.1  & 37.7  & 15.6 \\
          & min   & 17.0  & -1.1  & 15.7  & 4.9   & 1.6   & -2.7  & 1.2   & -1.4  & 0.0   & -15.3 & 2.7   & 0.0   & -2.1 \\
          & max   & 72.7  & 92.4  & 57.0  & 31.8  & 137.7 & 445.4 & 117.1 & 230.0 & 4119.5 & 1334.2 & 74.5  & 11696.6 & 204.3 \\
    \bottomrule
    \end{tabular}%
    % \begin{tablenotes}
    % \end{tablenotes}
  \label{tab:step1}%
\end{table}%

Table \ref{tab:step1} shows the statistical results after Step 1 extraction of longitudinal trajectory data. These results indicate the range of data collected in each dataset. For example, CATS UW Dataset, OpenACC ZalaZone Dataset, Waymo Perception Dataset, and Argoverse 2 Motion Forecasting Dataset (datasets 4, 8, 12, and 13) have a low average speed, suggesting that the scenarios in these four datasets are primarily low-speed environments. The CATS UW Dataset and OpenACC ZalaZone Dataset mainly test low-speed environments, while the Waymo Motion Dataset and Argoverse 2 Motion Forecasting Dataset are primarily collected in urban environments where the traffic conditions are complex, and AVs usually travel at low speeds. Additionally, there are some outliers, such as the maximum and minimum $a^{\mathrm{l}}$ and $a^{\mathrm{f}}$ in dataset 2, and the maximum $d$ in datasets 9 and 10, which would not occur in a normal driving process. We suppose that these data are caused by sensor errors and should be removed from the dataset.

\begin{table}[htbp]
  % \footnotesize
  \centering
  \caption{Statistical results of the data following Step 2 processing.}
    \begin{tabular}{llrrrrrrrrrrrrr}
    \toprule
    Label & Statistics & 1     & 2     & 3     & 4     & 5     & 6     & 7     & 8     & 9     & 10    & 11    & 12    & 13 \\
    \midrule
    \multirow{4}[2]{*}{$v^{\mathrm{l}}$} & mean  & 29.3  & 16.9  & 23.2  & 3.5   & 32.2  & 26.3  & 18.3  & 9.3   & 17.7  & 13.9  & 6.8   & 5.8   & 4.3 \\
          & std   & 1.5   & 7.2   & 0.7   & 0.5   & 6.6   & 9.5   & 4.7   & 3.9   & 10.4  & 10.2  & 7.4   & 6.2   & 4.8 \\
          & min   & 25.7  & 0.0   & 17.4  & 1.7   & 0.0   & 0.0   & 0.0   & -0.1  & 0.0   & 0.0   & -0.3  & 0.0   & 0.0 \\
          & max   & 34.8  & 26.7  & 24.7  & 4.0   & 40.1  & 40.8  & 33.8  & 27.5  & 37.4  & 33.4  & 28.5  & 30.7  & 22.9 \\
    \midrule
    \multirow{4}[2]{*}{$a^{\mathrm{l}}$} & mean  & 0.0   & 0.1   & 0.0   & 0.0   & 0.0   & 0.0   & 0.0   & 0.0   & 0.0   & 0.0   & 0.1   & 0.2   & 0.1 \\
          & std   & 0.3   & 0.6   & 0.2   & 0.0   & 0.6   & 0.6   & 0.5   & 0.4   & 1.0   & 1.0   & 0.7   & 0.9   & 1.1 \\
          & min   & -2.7  & -3.9  & -1.8  & 0.0   & -5.3  & -5.0  & -3.4  & -4.0  & -9.6  & -10.2 & -2.9  & -9.0  & -11.5 \\
          & max   & 3.8   & 5.4   & 1.6   & 0.2   & 5.9   & 5.8   & 3.2   & 4.0   & 9.6   & 10.2  & 3.5   & 9.3   & 11.8 \\
    \midrule
    \multirow{4}[2]{*}{$v^{\mathrm{f}}$} & mean  & 29.3  & 16.8  & 23.2  & 3.2   & 32.2  & 26.3  & 18.3  & 9.3   & 17.8  & 13.9  & 7.0   & 6.3   & 4.6 \\
          & std   & 1.7   & 7.4   & 0.9   & 0.7   & 6.9   & 9.6   & 4.9   & 4.0   & 10.2  & 10.0  & 7.1   & 6.3   & 5.0 \\
          & min   & 25.7  & 0.0   & 17.6  & 1.2   & 0.0   & 0.0   & 0.0   & 0.0   & 0.0   & 0.0   & -0.3  & 0.0   & 0.0 \\
          & max   & 34.9  & 28.6  & 25.4  & 4.2   & 41.2  & 42.8  & 34.5  & 27.5  & 34.3  & 31.9  & 29.6  & 36.8  & 24.0 \\
    \midrule
    \multirow{4}[2]{*}{$a^{\mathrm{f}}$} & mean  & 0.0   & 0.1   & 0.0   & 0.0   & 0.0   & 0.0   & 0.0   & 0.0   & 0.0   & 0.0   & 0.0   & 0.1   & 0.0 \\
          & std   & 0.2   & 0.7   & 0.2   & 0.1   & 0.6   & 0.6   & 0.5   & 0.4   & 0.6   & 0.6   & 0.7   & 2.6   & 1.0 \\
          & min   & -1.1  & -4.8  & -1.0  & -1.2  & -6.2  & -6.5  & -4.4  & -4.3  & -5.6  & -4.4  & -2.9  & -27.1 & -11.0 \\
          & max   & 0.6   & 4.2   & 2.0   & 1.0   & 6.1   & 6.5   & 3.4   & 4.3   & 6.1   & 3.3   & 2.9   & 27.3  & 11.2 \\
    \midrule
    \multirow{4}[2]{*}{$d$} & mean  & 36.2  & 31.9  & 37.9  & 13.8  & 42.6  & 36.3  & 25.3  & 20.6  & 26.7  & 20.1  & 20.5  & 12.6  & 14.1 \\
          & std   & 6.7   & 13.0  & 10.3  & 4.8   & 15.3  & 20.9  & 12.3  & 11.8  & 15.1  & 28.7  & 16.1  & 10.2  & 15.7 \\
          & min   & 17.0  & -1.1  & 15.7  & 5.0   & 1.6   & -2.7  & 1.2   & -1.2  & 0.0   & -13.8 & 2.7   & 0.0   & -2.1 \\
          & max   & 72.7  & 92.4  & 57.0  & 31.8  & 137.7 & 245.6 & 117.1 & 139.4 & 59.3  & 256.9 & 74.5  & 72.9  & 169.2 \\
    \bottomrule
    \end{tabular}%
  \label{tab:step2}%
\end{table}%

Due to the presence of outliers in the data obtained from Step 1, we proceeded with Step 2 general data cleaning, where the processed data are recorded in Table \ref{tab:step2}. After this step, the outliers initially found in Table \ref{tab:step1} have been removed. The data removal was carefully limited to a small quantity to ensure that the processing did not significantly influence the overall data distribution. Consequently, the mean and standard deviation remain largely consistent with those in Table \ref{tab:step1}. Nevertheless, some data that do not typically occur in the car-following scenario still exist in Table \ref{tab:step2}. For example, the minimum speeds of the OpenACC ZalaZone Dataset and Waymo Perception Dataset (datasets 8 and 11) are less than 0 m/s. The maximum acceleration and deceleration of the Ohio Two-vehicle Dataset and Waymo Motion Dataset (datasets 10 and 12) exceed 10 $\mathrm{m}/\mathrm{s}^2$. In the OpenACC Vicolungo Dataset and Ohio Two-vehicle Dataset (datasets 6 and 10), the maximum space gaps are around 250 m.

\begin{table}[htbp]
  % \footnotesize
  \centering
  \caption{Statistical results of the data following Step 3 processing.}
        \begin{tabular}{llrrrrrrrrrrrrr}
    \toprule
    Label & Statistics & 1     & 2     & 3     & 4     & 5     & 6     & 7     & 8     & 9     & 10    & 11    & 12    & 13 \\
    \midrule
    \multirow{4}[2]{*}{$v^{\mathrm{l}}$} & mean  & 29.3  & 17.8  & 23.2  & 3.5   & 32.6  & 27.3  & 18.6  & 9.9   & 20.3  & 20.7  & 10.8  & 6.6   & 7.9 \\
          & std   & 1.5   & 6.3   & 0.7   & 0.5   & 5.7   & 8.2   & 4.3   & 3.3   & 8.3   & 6.6   & 6.8   & 5.3   & 4.0 \\
          & min   & 25.7  & 0.1   & 17.4  & 1.7   & 0.2   & 0.1   & 0.1   & 0.1   & 0.1   & 0.1   & 0.1   & 0.1   & 0.1 \\
          & max   & 34.8  & 26.7  & 24.7  & 4.0   & 40.1  & 40.8  & 33.8  & 27.5  & 37.4  & 33.4  & 28.5  & 29.3  & 22.9 \\
    \midrule
    \multirow{4}[2]{*}{$a^{\mathrm{l}}$} & mean  & 0.0   & 0.1   & 0.0   & 0.0   & 0.0   & 0.0   & 0.0   & 0.0   & 0.0   & 0.0   & 0.1   & 0.1   & 0.1 \\
          & std   & 0.3   & 0.6   & 0.2   & 0.0   & 0.6   & 0.6   & 0.5   & 0.4   & 1.0   & 0.9   & 0.8   & 0.8   & 1.3 \\
          & min   & -2.3  & -3.9  & -1.8  & 0.0   & -5.0  & -4.8  & -3.4  & -4.0  & -5.0  & -4.9  & -2.9  & -5.0  & -5.0 \\
          & max   & 2.4   & 4.4   & 1.6   & 0.2   & 4.9   & 5.0   & 3.2   & 4.0   & 5.0   & 5.0   & 3.5   & 5.0   & 5.0 \\
    \midrule
    \multirow{4}[2]{*}{$v^{\mathrm{f}}$} & mean  & 29.3  & 17.6  & 23.2  & 3.2   & 32.6  & 27.3  & 18.6  & 9.9   & 20.3  & 20.4  & 10.7  & 7.0   & 8.3 \\
          & std   & 1.7   & 6.6   & 0.9   & 0.7   & 5.9   & 8.4   & 4.5   & 3.3   & 8.3   & 6.3   & 6.6   & 5.3   & 4.1 \\
          & min   & 25.7  & 0.1   & 17.6  & 1.2   & 0.1   & 0.1   & 0.1   & 0.1   & 0.1   & 0.1   & 0.1   & 0.1   & 0.1 \\
          & max   & 34.9  & 28.6  & 25.4  & 4.2   & 41.2  & 40.8  & 34.5  & 27.4  & 34.3  & 31.9  & 29.6  & 29.3  & 23.6 \\
    \midrule
    \multirow{4}[2]{*}{$a^{\mathrm{f}}$} & mean  & 0.0   & 0.1   & 0.0   & 0.0   & 0.0   & 0.0   & 0.0   & 0.0   & 0.0   & 0.0   & 0.1   & 0.1   & 0.1 \\
          & std   & 0.2   & 0.7   & 0.2   & 0.1   & 0.6   & 0.7   & 0.5   & 0.4   & 0.6   & 0.4   & 0.7   & 1.7   & 0.9 \\
          & min   & -1.1  & -4.8  & -1.0  & -1.2  & -4.7  & -5.0  & -4.4  & -4.3  & -5.0  & -4.0  & -2.9  & -5.0  & -5.0 \\
          & max   & 0.6   & 4.2   & 2.0   & 1.0   & 4.9   & 4.9   & 3.2   & 4.3   & 4.6   & 2.9   & 2.8   & 5.0   & 5.0 \\
    \midrule
    \multirow{4}[2]{*}{$d$} & mean  & 36.2  & 33.4  & 37.9  & 13.8  & 42.7  & 36.1  & 25.6  & 21.5  & 31.6  & 36.7  & 28.0  & 12.1  & 16.9 \\
          & std   & 6.8   & 11.6  & 10.3  & 4.8   & 13.7  & 16.4  & 12.1  & 11.0  & 11.8  & 21.6  & 15.0  & 6.8   & 13.4 \\
          & min   & 17.0  & 1.4   & 15.7  & 5.0   & 1.6   & 0.0   & 1.5   & 0.0   & 0.0   & 0.0   & 3.6   & 0.5   & 0.0 \\
          & max   & 72.7  & 92.4  & 57.0  & 31.8  & 119.6 & 119.9 & 117.1 & 120.0 & 59.3  & 120.0 & 74.5  & 68.6  & 120.0 \\
    \bottomrule
    \end{tabular}%
  \label{tab:step3}%
\end{table}%

Table \ref{tab:step3} shows the statistical results after Step 3 data-specific cleaning. Following the removal of non-car-following scenario data, the data in Table \ref{tab:step2} have been adjusted to normal ranges. It is evident that the average speeds across all datasets have increased, especially in the Ohio Single-vehicle Dataset, Ohio Two-vehicle Dataset, Waymo Perception Dataset, and Argoverse 2 Motion Forecasting Dataset (datasets 9, 10, 11, and 13). This suggests the presence of numerous stationary scenarios within these datasets. The average spatial gaps in the Ohio Single-vehicle Dataset, Ohio Two-vehicle Dataset, and Waymo Perception Dataset (datasets 9, 10, and 11) have also significantly increased, indicating that these datasets initially contained many instances of small gaps.

\begin{figure}
    \centering
    \includegraphics[width=1\linewidth]{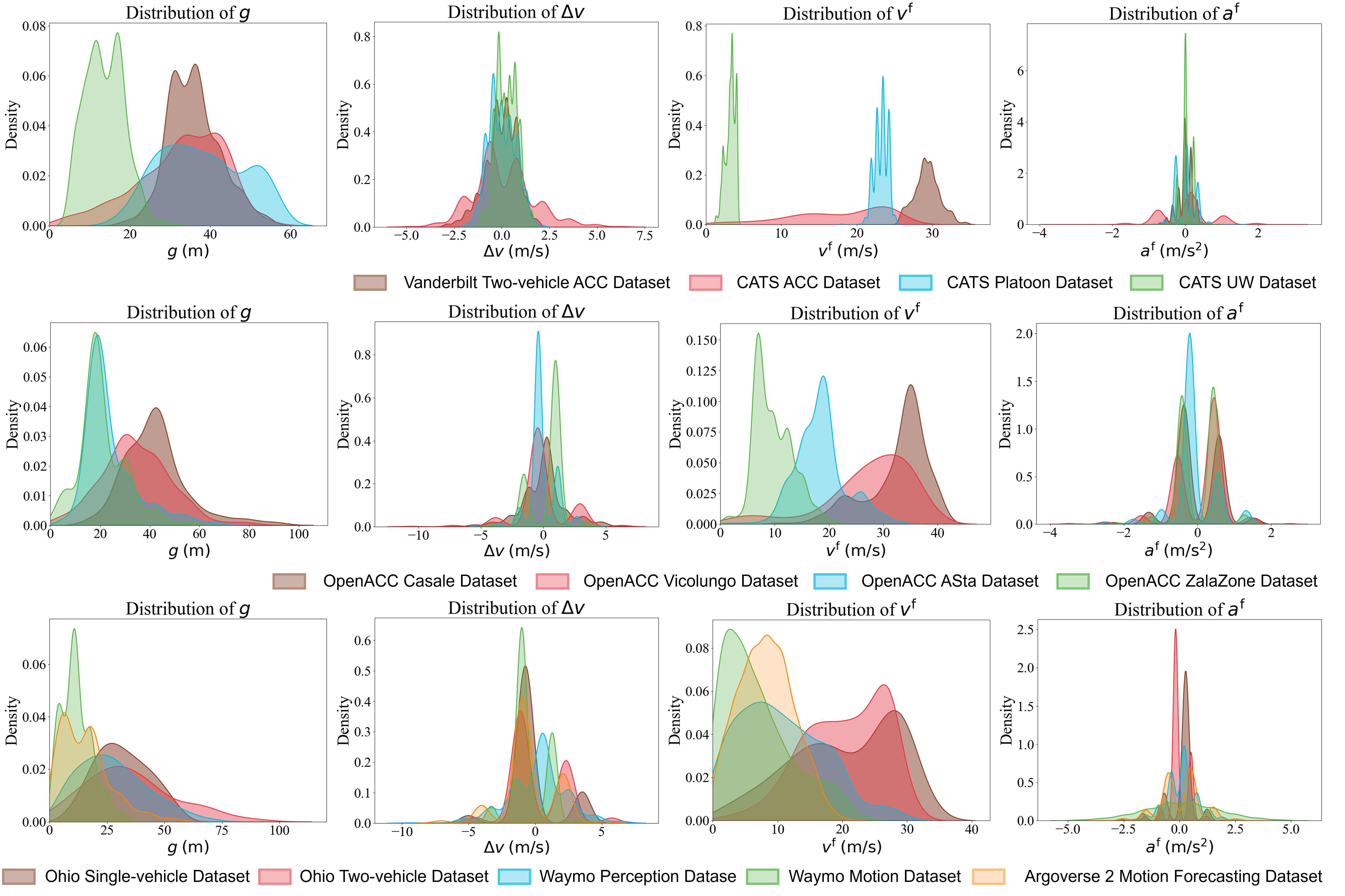}
    \caption{Probability distributions of $g$, $\Delta v$, $v^{\mathrm{f}}$, and $a^{\mathrm{f}}$.}
    \label{fig:statistics}
\end{figure}

Figure \ref{fig:statistics} displays the statistical distributions of key variables in car-following behavior analysis, including spatial gap, relative speed, FAV speed, and FAV acceleration, for the final data after completing the three-step data processing. From Figure \ref{fig:statistics}, we can suppose the testing scenarios of the data, where some of them are indicated in Table \ref{tab:summary}. For example, the CATS UW Dataset was collected in a low-speed environment, while the OpenACC Casale Dataset was collected in a high-speed environment. Additionally, the form of the car-following model can be analyzed by observing the shape of the distribution. For example, the distributions of $a^{\mathrm{f}}$ from the four datasets in the OpenACC Database clearly show two distinct peaks, one on the left and one on the right of zero, possibly representing vehicles' behaviors in accelerating and decelerating are piecewise.

%The Data Records section should be used to explain each data record associated with this work, including the repository where this information is stored, and to provide an overview of the data files and their formats. Each external data record should be cited numerically in the text of this section, for example \cite{Hao:gidmaps:2014}, and included in the main reference list as described below. A data citation should also be placed in the subsection of the Methods containing the data-collection or analytical procedure(s) used to derive the corresponding record. Providing a direct link to the dataset may also be helpful to readers (\hyperlink{https://doi.org/10.6084/m9.figshare.853801}{https://doi.org/10.6084/m9.figshare.853801}).

%Tables should be used to support the data records, and should clearly indicate the samples and subjects (study inputs), their provenance, and the experimental manipulations performed on each (please see 'Tables' below). They should also specify the data output resulting from each data-collection or analytical step, should these form part of the archived record.

\section*{Technical Validation}

%This section presents any experiments or analyses that are needed to support the technical quality of the dataset. This section may be supported by figures and tables, as needed. This is a required section; authors must present information justifying the reliability of their data.

In this section, we validate the processed unified trajectory dataset through three aspects. First, we introduce the data collection methods we used in the CATS Open Datasets. Then, we analyze the performance of the car-following trajectory dataset through four metrics. Finally, we analyze the relationships between the variables in the car-following model.

\subsection*{Data Collection}

First, we introduce the AV platform developed by the CATS Lab as a reference solution for future AV trajectory data collection. The CATS Lab has developed a complete AV platform, which has been set up in two lab-owned Lincoln MKZ. The platform is built upon the Robot Operating System (ROS), which provides a robust framework for parallel computing and is particularly well-suited for robotics and autonomous applications. In addition, the platform allows for direct electronic control over the vehicle's functions by integrating the Drive-By-Wire (DBW) system. 

\begin{figure}
    \centering
    \includegraphics[width=1\linewidth]{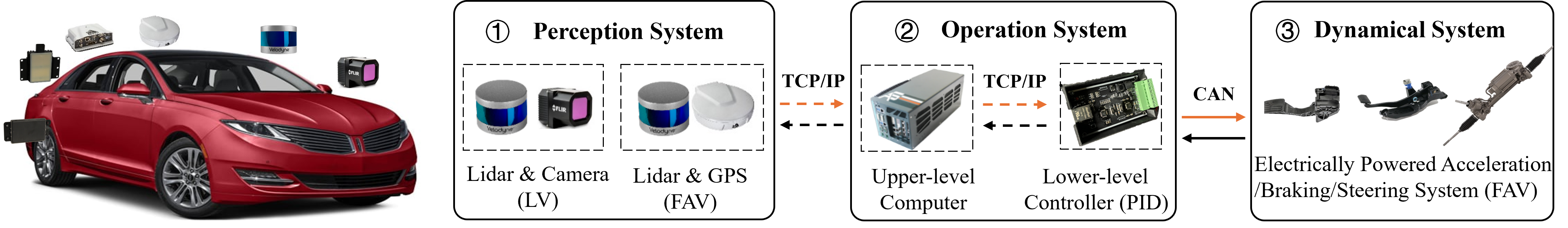}
    \caption{AV platform developed by the CATS Lab.}
    \label{fig:platform}
\end{figure}

The developed system includes the perception system, operation system, and dynamical system, shown in Figure \ref{fig:platform}. The perception system comprises advanced sensing technologies such as LiDAR and cameras that provide real-time data on the vehicle's surroundings. LiDAR and GPS navigation units offer high-precision location tracking capabilities. The operation system is a hierarchical structure with an upper-level computer and a lower-level control. This system utilizes the Transmission Control Protocol/Internet Protocol (TCP/IP) protocol for networking and communication, interlinking with the Controller Area Network (CAN) for vehicle control. The dynamical system features an electrically powered acceleration/braking/steering system to manipulate the vehicle's longitudinal and lateral motions. The final output utilizes the CAN to transit the signals for brake, throttle, and steering angle.

Data collection is accomplished by the perception system. The perception system obtains precise vehicle trajectory data collected from various sensors, including LiDAR, GPS, and cameras. The early data collected by CATS Lab, including the CATS ACC Dataset and CATS Platoon Dataset, primarily used the Ublox GPS system to gather real-time GPS positions and speeds. The real-time vehicle-following spacing between the two vehicles could be obtained by the distance between the GPS positions. Preliminary testing indicated that the GPS receivers had a position accuracy of 0.26 m and a speed accuracy of 0.089 m/s. Due to low precision and the data packet loss during transmission, in the CATS UW Dataset, we utilized LiDAR for data collection. To achieve higher data precision, a feasible approach is to design algorithms that integrate data from multiple sensors. The Waymo Open Dataset and Argoverse Dataset have already employed similar technologies. Therefore, we recommend that researchers integrate multiple sensors in future data collection to obtain high-precision data.

\subsection*{Performance Measurement}

Next, we analyze how each dataset impacts road traffic performance in four metrics, i.e., safety, mobility, stability, and sustainability. We first define the measurements of these four metrics.

The safety of FAV in car-following behavior is measured by the Time-To-Collision \cite{minderhoud2001extended} (TTC) to represent the risk or proximity of a vehicle to a potential collision. The TTC at time $t$ is defined as the time that remains until a collision between two vehicles would have occurred if the collision course and speed difference were maintained. The higher the TTC is, the more safe a situation is, and vice versa \cite{minderhoud2001extended}. The TTC in trajectory $i$ at time $t$ can be calculated as follows:
\begin{align}
    TTC_{it}=\frac{g_{it}}{v^{\mathrm{f}}_{it}-v^{\mathrm{l}}_{it}}, \quad v^{\mathrm{f}}_{it} > v^{\mathrm{l}}_{it}.
\end{align}

The mobility is measured by the time headway, defined by the time difference between consecutive arrival instants of two vehicles passing a certain detector site on the same lane \cite{ha2012time}. The time headway is considered a direct measure of road capacity. A short time headway will increase road capacity and thus increase mobility, and vice versa \cite{li2022trade}.
% Study has shown that, for human drivers, the lower limit of the time headway is about 1 second \cite{ayres2001preferred}. 
The time headway in trajectory $i$ at time $t$ can be calculated as follows:
\begin{align}
    \tau_{it}=\frac{h_{it}}{v^{\mathrm{f}}_{it}}.
\end{align}

The stability is measured by the squared error of acceleration at time $t$. Larger variations in acceleration are considered a lack of smoothest to indicate discomfort and potential safety risks. The larger the squared error of acceleration is, the lower traffic stability,  and vice versa. It can be calculated as follows:
\begin{align}
    \alpha_{it}=(a_{it}-\frac{\sum_{t'\in \mathcal{T}_i} a_{it'}}{T_i})^2.
\end{align}

The sustainability is measured by the fuel consumption rate of the FAV. We utilize the average value of four classical vehicle fuel consumption models to measure the fuel consumption, including the Virginia Tech Microscopic (VT-Micro) model \cite{zegeye2013integrated}, Microscopic Emission and Fuel consumption (MEF) model \cite{lei2010microscopic}, Vehicle Specific Power (VSP) model \cite{duarte2015establishing}, and Australian Road Research Board (ARRB) model \cite{akcelik1989efficiency,knoop2019platoon}. We use $F^{\mathrm{VTM}}_{it}, F^{\mathrm{MEF}}_{it}, F^{\mathrm{VSP}}_{it}, F^{\mathrm{ARRB}}_{it}$ to represent the fuel consumption calculated by the four models in trajectory $i$ at time $t\in \mathcal{T}_i$, respectively. To simplify the notation, in equations \eqref{equ:VTM}-\eqref{equ:ARRM}, we use $v_{it}$ and $a_{it}$ to represent the velocity and acceleration of the vehicle in trajectory $i$ at time $t\in \mathcal{T}_i$. The expression of the four models is shown as follows:
\begin{itemize}
    \item VT-Micro model:
    \begin{align}
        F^{\mathrm{VTM}}_{it}=\exp \left(\sum_{n_1=0}^3 \sum_{n_2=0}^3 K_{n_1 n_2}\left(v_{it}\right)^{n_1}\left(a_{it}\right)^{n_2}\right) \label{equ:VTM}
    \end{align}
    where $n_1, n_2$ are the power indexes and $K_{n_1, n_2}$ are constant coefficients, which are available in Appendix A.
    \item MEF model:
    \begin{align}
        F^{\mathrm{MEF}}_{it}=\exp \left(\sum_{n_1=0}^3 \sum_{n_2=0}^3 K_{n_1 n_2}\left(v_t\right)^{n_1}\left(\beta \cdot a_{it}+(1-\beta) \sum_{t'=1}^T a_{i(t-t')} / T \right)^{n_2}\right)
    \end{align}
    where $\beta=0.5$ is the acceleration impact factor, and $T=9$ is the number of historical data considered in the model.
    \item VSP model:
    \begin{align}
         & VSP_{it}=v_{it}\left(1.1 a_{it}+9.81 \delta+0.132\right)+3.02E-4 \cdot v_{it}^3 \\ 
         & F^{\mathrm{VSP}}_{it}=\left\{\begin{array}{cc}
                f_1, & \text {if } VSP_{it}<-10 \\
                f_2 VSP_{it}^2+f_3 VSP_{it}+f_4, & \text {if } -10 \leq VSP_{it} \leq 10 \\
                f_5 VSP_{it}+f_6, & \text {if } VSP_{it} \geq 10
            \end{array}\right.
    \end{align}
    where $\delta$ denotes the road grade that is set to 0 in this paper since we assume the road grade can be neglected in most experiment sites, $VSP_{it}$ is the vehicle specific power in trajectory $i$ at time $t\in \mathcal{T}_i$. Parameters $f_1 = 2.48E-03, f_2 = 1.98E-03, f_3 = 3.97E-02, f_4 = 2.01E-01, f_5 = 7.93E-02, f_6 = 2.48E-03$, where the notation 'E' represents exponentiation in scientific notation.
    \item ARRB model:
    \begin{align}
        F^{\mathrm{ARRB}}_{it}=\gamma_1+\gamma_2 v_{it}+\gamma_3 v_{it}^2+\gamma_4 v_{it}^3+\gamma_5 v_{it} \cdot a_{it}+\gamma_6 v_{it}\left(\max \left(0, a_{it}\right)^2\right) \label{equ:ARRM}
    \end{align}
    where parameters $ \gamma_1 = 0.666, \gamma_2 = 0.019, \gamma_3 = 0.001, \gamma_4 = 0.0005, \gamma_5 = 0.122$, and $\gamma_6 = 0.793$.
\end{itemize}
In equations \eqref{equ:VTM}-\eqref{equ:ARRM}, the units for $v_{it}$ and $a_{it}$ are m/s and m/$\mathrm{s}^2$, respectively. The units for $F^{\mathrm{VTM}}_{it}, F^{\mathrm{MEF}}_{it}, F^{\mathrm{VSP}}_{it}, F^{\mathrm{ARRB}}_{it}$ are L/s, L/s, g/s, and ml/s, respectively. To perform unit conversions, we assume the density of fuel is 800 g/L (the density of gasoline at room temperature is approximately 720 to 775 g/L, and diesel is about 830 to 850 g/L). Therefore, the total energy consumption equation is formulated as:
\begin{align}
    F^{\mathrm{all}}_{it}= \left(F^{\mathrm{VTM}}_{it} + F^{\mathrm{MEF}}_{it} + F^{\mathrm{VSP}}_{it} / 800 + F^{\mathrm{ARRB}}_{it} / 1000\right) / 4
\end{align}

\begin{figure}
    \centering
    \includegraphics[width=1\linewidth]{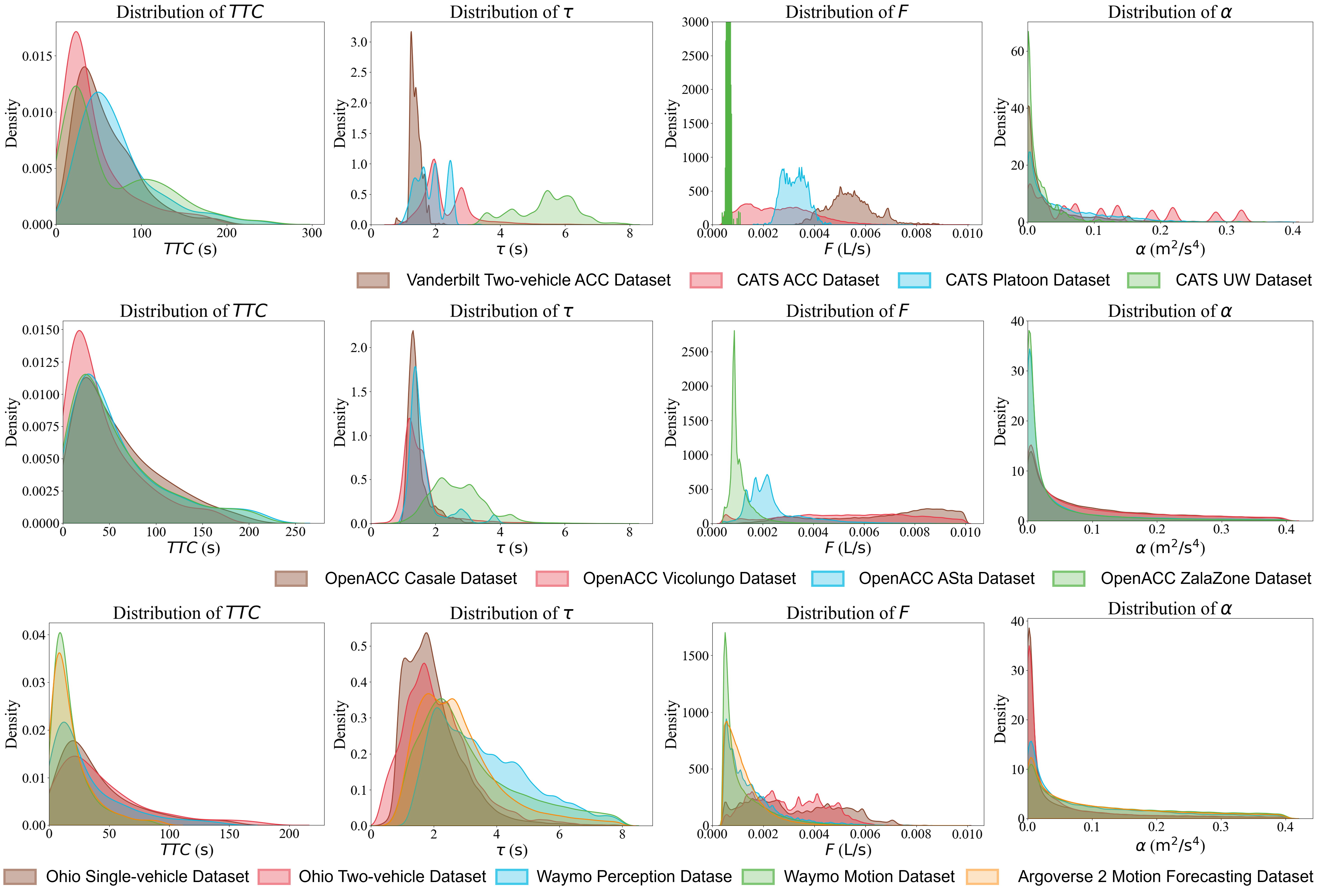}
    \caption{Distribution of performance metrics of safety, mobility, stability, and sustainability.}
    \label{fig:performance}
\end{figure}

Figure \ref{fig:performance} displays the distributions of the four indicators across all datasets. The distribution of $TTC$ is predominantly left-skewed, with peak values below 50 s. Notably, the Waymo Motion Dataset and Argoverse 2 Motion Forecasting Dataset exhibit the smallest peaks around 10 s, and their probability density is concentrated in the lower TTC range. Given that a smaller TTC indicates greater risk, this distribution suggests a higher risk associated with AVs in complex traffic environments.

The distribution of $\tau$ is primarily concentrated between 1-5 s. The CATS Platoon Dataset shows multiple peaks due to testing with four levels of time headway settings. In contrast, the CATS UW Dataset exhibits a larger $\tau$, indicating poorer mobility of its AV's car-following model compared to the smaller $\tau$ observed in the OpenACC Database and Vanderbilt ACC Dataset. This difference may also caused by the different experimental settings. The CATS UW Dataset was tested in a low-speed environment with a minimum safety distance, leading to a larger $\tau$, while the other two datasets were tested at higher speeds.

Regarding the distribution of $F$, some datasets, including the CATS UW Dataset, OpenACC ZalaZone Dataset, Waymo Motion Dataset, and Argoverse 2 Motion Forecasting Dataset, predominantly have $F$ below 0.001 L/s. Conversely, the Vanderbilt ACC Dataset, OpenACC Casale Dataset, and OpenACC Vicolungo Dataset exhibit higher $F$, averaging over 0.005 L/s. The variance in $F$ across these datasets can be explained by the positive correlation between energy consumption and speed. According to Table \ref{tab:step3}, the datasets with the lower $F$ correspond to those with the lowest average speeds, while the datasets with higher $F$ have the highest average speeds. Additionally, differences in the vehicle car-following models' energy efficiency across the datasets also influence it. This result reflects the distinct energy consumption characteristics associated with the vehicles in each dataset.

Lastly, the distribution of $\alpha$ is centered near zero for most datasets, with the probability density decreasing as $\alpha$ increases. Among them, the CATS ACC Dataset shows fluctuations in its distribution. The reason is that the acceleration precision of the original data is limited to one decimal place, resulting in insufficient data accuracy. This causes $\alpha$ to be concentrated in a few areas. Additionally, the CATS Platoon Dataset, OpenACC Vicolungo Dataset, and Waymo Perception Dataset exhibit slightly higher densities at larger $\alpha$, indicating relatively poorer vehicle stability in these datasets.

Overall, the results presented in Figure \ref{fig:performance} are consistent with the literature. For example, $TTC$ in most datasets is concentrated below 50 seconds, and the time headway $\tau$ is mainly between 1 and 2 seconds. This demonstrates that the Ultra-AV dataset can be used for research on AV behavior analysis. The four metrics used in this paper can be utilized to evaluate the trajectory datasets collected in the future, and also serve as standards for the development of car-following models. Additionally, we observe the relationships between different metrics in Figure \ref{fig:performance}: 1. $\tau$ and $TTC$ show a negative correlation. 2. $\tau$ and $F$ show a negative correlation. 3. $\tau$ and $\alpha$ show a negative correlation. This indicates that improving some metrics might lead to worse outcomes in others. For example, increasing mobility by reducing $\tau$ might reduce the spatial gap between vehicles, thereby reducing $TTC$ and compromising vehicle safety. A future research direction is to develop models that make trade-offs between these four metrics to achieve overall optimal.

\subsection*{Car-following Model Development}

Although the trajectory data from FAV offers valuable insights into their performance, this data may stem from a limited set of conditions. These derived performance metrics may not reflect the full range of driving scenarios. Thus, the development of accurate and robust car-following models for simulation across a broader range of scenarios is advantageous. Researchers are trying to develop car-following models, including the linear ACC model \cite{ma2022string}, nonlinear intelligent driver model \cite{treiber2000congested}, or data-driven models \cite{zhu2018human}. No matter what the model structure is, they all adapted $a^{\mathrm{f}}_{it}$ as the output and $d_{it}$, $v^{\mathrm{f}}_{it}$, and $\Delta v_{it}$ as the input in these models. Thus, we analyze the relationship between output acceleration and input three variables with scatter plots and correlation analysis.

\begin{figure}
    \centering
    \includegraphics[width=0.9\linewidth]{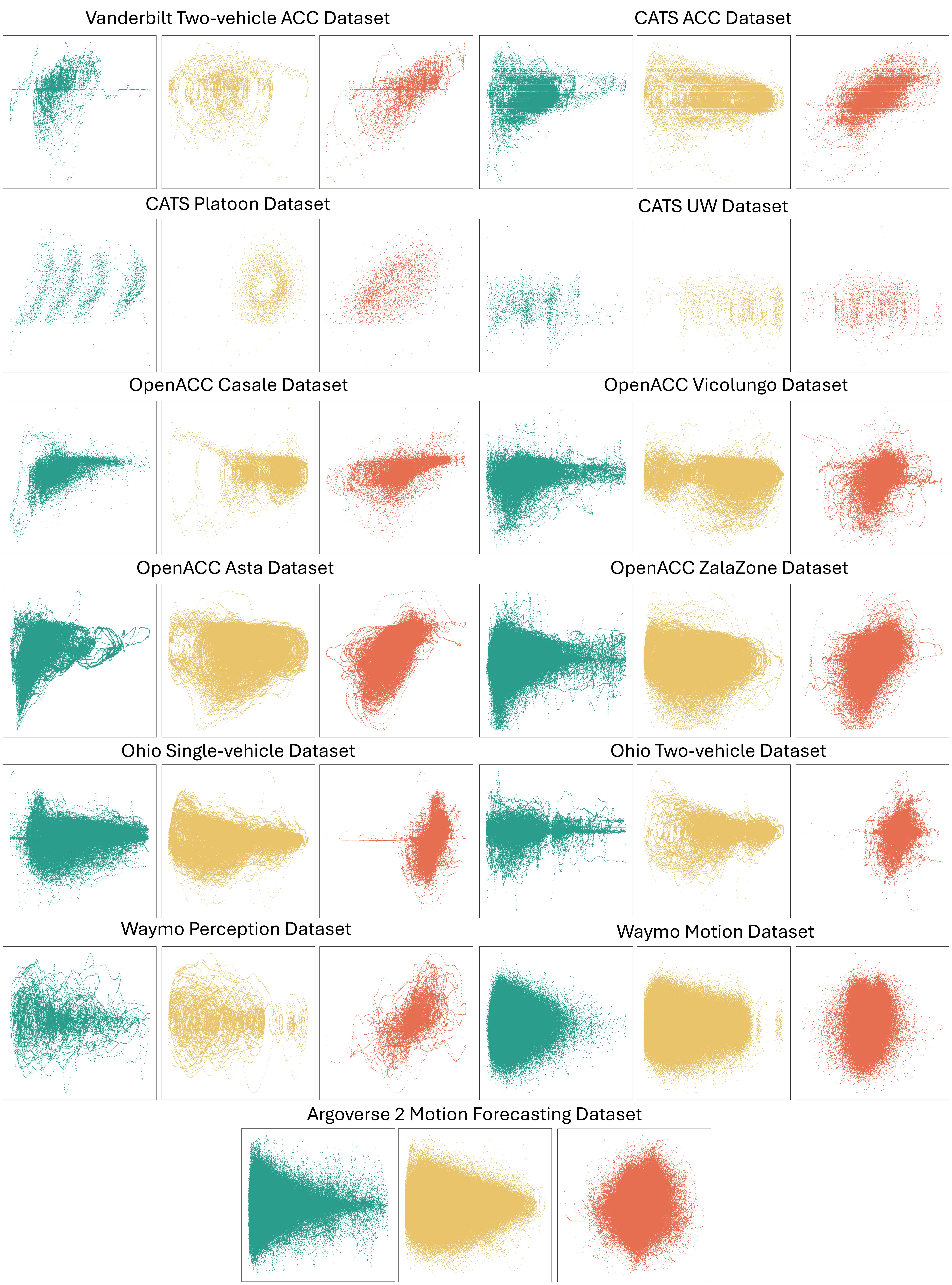}
    \caption{The scatter plots between $a^{\mathrm{f}}$ and $d$, $v^{\mathrm{f}}$, and $\Delta v$. The y-axis represents $a^{\mathrm{f}}$, and the x-axis with green, yellow, and red colors represent $d$, $v^{\mathrm{f}}$, and $\Delta v$, respectively.}
    \label{fig:scatter}
\end{figure}

Figure \ref{fig:scatter} displays the relationship between output acceleration and input three variables. To accurately depict the relationships among the variables, the moving average method with a window length of three was applied to smooth the $a^{\mathrm{f}}_{it}$ data before plotting. In Figure \ref{fig:scatter}, a notably nonlinear positive correlation between $a^{\mathrm{f}}_{it}$ and $d_{it}$ is evident, particularly in the datasets from the OpenACC Database, which a logarithmic curve can characterize. Besides, most datasets show a linear positive correlation between $a^{\mathrm{f}}_{it}$ and $\Delta v_{it}$. In contrast, there is no clear relationship between $a^{\mathrm{f}}_{it}$ and $v_{it}$. The unclear relationship is due to $a^{\mathrm{f}}_{it}$ being influenced by the other three variables collectively, and it cannot be directly reflected by a single variable. Therefore, a deep analysis of the relationships among variables in car-following behavior is necessary, as well as the development of specific car-following models. Moreover, car-following models may vary across different vehicle types, and more detailed studies should analyze the same vehicle. This paper will not delve into specific models, but researchers can refer to the latest reviews in this field \cite{brackstone1999car, wang2023car} to conduct studies using the provided data.

\begin{table}[htbp]
  \centering
  \caption{Correlation coefficients between $a^{\mathrm{f}}$ and $d$, $v^{\mathrm{f}}$, and $\Delta v$.}
        \begin{tabular}{crrrrrr}
    \toprule
    \multirow{2}[4]{*}{Dataset} & \multicolumn{3}{c}{Pearson} & \multicolumn{3}{c}{Spearman} \\
    \cmidrule{2-7}          & \multicolumn{1}{c}{$d$} & \multicolumn{1}{c}{$v^{\mathrm{f}}$} & \multicolumn{1}{c}{$\Delta v$} & \multicolumn{1}{c}{$d$} & \multicolumn{1}{c}{$v^{\mathrm{f}}$} & \multicolumn{1}{c}{$\Delta v$} \\
    \midrule
    1     & 0.3954 & 0.0040 & 0.6606 & 0.4987 & 0.0323 & 0.6518 \\
    2     & 0.1557 & -0.1246 & 0.5803 & 0.1828 & -0.1005 & 0.5644 \\
    3     & 0.1215 & -0.1255 & 0.7076 & 0.1609 & -0.1136 & 0.7317 \\
    4     & 0.0027 & -0.1708 & 0.0163 & 0.0160 & -0.1732 & 0.0346 \\
    5     & 0.3373 & -0.0007 & 0.3512 & 0.3450 & -0.0264 & 0.3577 \\
    6     & 0.2193 & -0.0352 & 0.4052 & 0.2474 & -0.0673 & 0.4838 \\
    7     & 0.2265 & -0.0062 & 0.6369 & 0.3583 & -0.0550 & 0.5575 \\
    8     & 0.1377 & -0.0124 & 0.5767 & 0.1559 & 0.0049 & 0.5798 \\
    9     & 0.1289 & -0.0412 & 0.4792 & 0.1859 & -0.1359 & 0.5228 \\
    10    & 0.1009 & 0.0092 & 0.2333 & 0.0932 & -0.0728 & 0.2263 \\
    11    & -0.0763 & -0.0925 & 0.5466 & -0.0835 & -0.0941 & 0.5660 \\
    12    & 0.0102 & -0.0219 & 0.1080 & 0.0210 & -0.0150 & 0.1061 \\
    13    & 0.0043 & -0.0647 & 0.2764 & 0.0446 & -0.0746 & 0.3087 \\
    \bottomrule
    \end{tabular}%
  \label{tab:correlation}%
\end{table}%

The relationship between $a^{\mathrm{f}}$ and three variables can also be revealed with correlation coefficients. Table \ref{tab:correlation} shows the Pearson and Spearman correlation coefficients between $a^{\mathrm{f}}$ and $d$, $v^{\mathrm{f}}$, and $\Delta v$ for all datasets. According to Table \ref{tab:correlation}, $a^{\mathrm{f}}$ has a positive correlation with $d$ and $\Delta v$, and a negative correlation with $v^{\mathrm{f}}$. This result is similar to the experience: when $d$ is large, the FAV accelerates to close the gap; when $v^{\mathrm{f}}$ is large, the FAV slows down to stabilize at a following speed; and when $\Delta v$ is positive, the FAV decelerates to match the LV speed. The correlation coefficient for $d$ generally ranges from 0 to 0.4, indicating a weak correlation. For $v^{\mathrm{f}}$, the coefficient ranges from -0.2 to 0, indicating no correlation For $\Delta v$, the coefficient is above 0.5, indicating a strong correlation. Since Pearson and Spearman correlation coefficients only reflect the linear and monotonic relationships between variables, we suppose that in the car-following model, the influence of $v^{\mathrm{f}}$ on $a^{\mathrm{f}}$ is nonlinear.  This insight inspires us to develop piecewise linear or nonlinear car-following models.

From the analysis of the scatter plots and correlation analysis, it is clear that the Ultra-AV dataset reflects certain relationships between acceleration and other variables. The results show that the relationships depicted in the dataset are similar to those discussed in the literature and experience, validating that our dataset can be utilized for the development of car-following models. The analysis also indicates that there is a certain nonlinearity in the relationships between variables and acceleration, particularly with $v^{\mathrm{f}}$. This inspires researchers to consider the nonlinear relationship between $v^{\mathrm{f}}$ and $a^{\mathrm{f}}$ in future model development.

\section*{Usage Notes}

In this study, we reviewed a series of AV trajectory datasets and extracted the Ultra-AV dataset. This dataset includes two sub-datasets: the longitudinal trajectory dataset and the car-following trajectory dataset, corresponding to data after Step 2 general data cleaning, and Step 3 data-specific cleaning, respectively.

Contrary to the existing AV perception datasets discussed in the literature, this dataset offers AV trajectory data that enhances the analysis of microscopic longitudinal AV behaviors. Few existing trajectory datasets were stored in their respective formats and often fall short in terms of refinement, reliability, and completeness, which limits their widespread use and comparison. Therefore, this paper reviewed most of the trajectory datasets, transformed all datasets into a unified format through data processing, and performed cleaning and refinement.

Moreover, we analyzed the data using multiple methods to validate its reliability. We first analyzed the impact of the data collection methods. We recommend researchers integrate data from multiple sensors such as GPS, LiDAR, and cameras, to enhance the precision of trajectory data and preserve essential information for AV behavior analysis. Secondly, we summarized the performance of each dataset in safety, mobility, sustainability, and stability under four metrics widely used in the literature. These four metrics can be utilized to evaluate the trajectory datasets collected in the future. The results show that there are some relationships between different metrics, and the trade-offs between metrics need to be considered during model design. Finally, we analyzed the relationships between variables in the car-following model through correlation coefficients and scatter plots, suggesting that researchers focus on the nonlinear impact of FAV speed on acceleration in model development.

In this study, we focus on the longitudinal behaviors of AVs, not considering lateral behaviors, such as lane-changing. This is due to the scarcity of trajectory datasets that include AV lateral behaviors, and the corresponding analyses and processing are more complex. However, we noted that some trajectory datasets, such as the Central Ohio ACC Dataset, Waymo Open Dataset, and Argoverse 2 Motion Forecasting Dataset, contain lane-changing behaviors. To fully understand AV behaviors, it is essential to consider both longitudinal and lateral behaviors. Consequently, developing an AV behavior model that balances the four metrics is important. Moreover, how to utilize the results of the behavior model to improve the AV would enhance the transportation system.

%The Usage Notes should contain brief instructions to assist other researchers with the reuse of the data. This may include a discussion of software packages that are suitable for analyzing the assay data files, suggested downstream processing steps (e.g. normalization, etc.), or tips for integrating or comparing the data records with other datasets. Authors are encouraged to provide code, programs, or data-processing workflows if they may help others understand or use the data. Please see our code availability policy for advice on supplying custom code alongside Data Descriptor manuscripts.

%For studies involving privacy or safety controls on public access to the data, this section should describe in detail these controls, including how authors can apply them to access the data, what criteria will be used to determine who may access the data, and any limitations on data use. 

\section*{Code availability}

The Ultra-AV dataset and the code for data extraction and analysis have been documented and made accessible on \href{https://github.com/CATS-Lab/Filed-Experiment-Data-Unified-AV-Trajectory-Dataset}{https://github.com/CATS-Lab/Filed-Experiment-Data-Unified-AV-Trajectory-Dataset}. The detailed contents include:

\begin{itemize}
    \item Readme.md: A general description of the raw data for each dataset.
    \item Main.py: The main function calls data processing and analysis functions for each dataset.
    \item trajectory\_extraction.py: Code used in Step 1 to extract AV longitudinal trajectories.
    \item data\_transformation.py: Code used in Step 1 to convert all datasets to a unified format.
    \item data\_cleaning.py: Code used in Steps 2 and 3 for data cleaning.
    \item data\_analysis.py: Code used to analyze data statistics, plot traffic performance of datasets, and plot scatter plots.
    \item model\_calibration.py: An example tool to use the processed data to calibrate a linear car-following model.
\end{itemize}

We also recommend using other software packages such as R to effectively analyze the trajectory data. These tools are well-suited for handling the dataset's format.

Data usage is restricted to research purposes only. Any commercial exploitation of the data requires separate approval and possibly additional agreements.  

\bibliography{sample}

%\noindent LaTeX formats citations and references automatically using the bibliography records in your .bib file, which you can edit via the project menu. Use the cite command for an inline citation, e.g. \cite{Kaufman2020, Figueredo:2009dg, Babichev2002, behringer2014manipulating}. For data citations of datasets uploaded to e.g. \emph{figshare}, please use the \verb|howpublished| option in the bib entry to specify the platform and the link, as in the \verb|Hao:gidmaps:2014| example in the sample bibliography file. For journal articles, DOIs should be included for works in press that do not yet have volume or page numbers. For other journal articles, DOIs should be included uniformly for all articles or not at all. We recommend that you encode all DOIs in your bibtex database as full URLs, e.g. https://doi.org/10.1007/s12110-009-9068-2.

\section*{Acknowledgements} (not compulsory)

This research is sponsored by National Science Foundation, USA through Grants CMMI \#1932452 and CMMI \#2343167.

\section*{Author contributions statement}

H.Z. conducted the experiments and analyzed the results, K.M. conceived the experiment, S.L. conducted the experiments, X.L. and X.Q. revised the paper. All authors reviewed the manuscript. 

\section*{Competing interests} 

The authors declare no competing interests. 
%The corresponding author is responsible for providing a \href{https://www.nature.com/sdata/policies/editorial-and-publishing-policies#competing}{competing interests statement} on behalf of all authors of the paper. This statement must be included in the submitted article file.

\section*{Figures \& Tables}

%Figures, tables, and their legends, should be included at the end of the document. Figures and tables can be referenced in \LaTeX{} using the ref command, e.g. Figure \ref{fig:stream} and Table \ref{tab:example}. 

%Authors are encouraged to provide one or more tables that provide basic information on the main ‘inputs’ to the study (e.g. samples, participants, or information sources) and the main data outputs of the study. Tables in the manuscript should generally not be used to present primary data (i.e. measurements). Tables containing primary data should be submitted to an appropriate data repository.

%Tables may be provided within the \LaTeX{} document or as separate files (tab-delimited text or Excel files). Legends, where needed, should be included here. Generally, a Data Descriptor should have fewer than ten Tables, but more may be allowed when needed. Tables may be of any size, but only Tables which fit onto a single printed page will be included in the PDF version of the article (up to a maximum of three). 

%Due to typesetting constraints, tables that do not fit onto a single A4 page cannot be included in the PDF version of the article and will be made available in the online version only. Any such tables must be labelled in the text as ‘Online-only’ tables and numbered separately from the main table list e.g. ‘Table 1, Table 2, Online-only Table 1’ etc.

\appendix
\section{Coefficients of the VT-Micro model and MEF model}
\begin{table}[ht]
    \centering
    \caption{Coefficients of the VT-Micro model and MEF model.}
    \begin{tabular}{cllll}
    \hline
    $K_{n_1n_2}$ & $n_2 = 0$ & $n_2 = 1$ & $n_2 = 2$ & $n_2 = 3$ \\ 
    \hline
    $n_1 = 0$ & -7.537 & 0.4438 & 0.1716 & -0.0420 \\
    $n_1 = 1$ & 0.0973 & 0.0518 & 0.0029 & 0.0071 \\
    $n_1 = 2$ & -0.003 & -7.42E-04 & 1.09E-04 & 1.16E-04 \\
    $n_1 = 3$ & 5.3E-05 & 6E-06 & -1E-05 & -6E-06 \\
    \hline
    \end{tabular}
    \label{tab:VTM}
\end{table}

\end{document}